\documentclass[journal]{IEEEtran}

\usepackage{amssymb}
\usepackage[cmex10]{amsmath}
\usepackage{bm}
\usepackage{algorithm}
\usepackage{algorithmic}
\usepackage{multirow}
\usepackage{xcolor}
\usepackage{mathrsfs}
\usepackage{graphicx,color}
\graphicspath{{figures/}}

\usepackage{subfig}

\ifCLASSOPTIONcompsoc
  \usepackage[nocompress]{cite}
\else
  \usepackage{cite}
\fi
\ifCLASSINFOpdf
\else
\fi
\usepackage{url}
\begin{document}
%
\title{Directional Statistics-based Deep Metric Learning for Image Classification and Retrieval}
%
%
%
%

\author{Xuefei~Zhe,~
        Shifeng~Chen,~
        and~Hong~Yan,~\IEEEmembership{Fellow,~IEEE}

\thanks{Xuefei Zhe and Hong Yan are with the Department of Electronic Engineering, City University of Hong Kong, Hong Kong.}
\thanks{Shifeng Chen is with Shenzhen Institutes of Advanced Technology, CAS, China.}
\thanks{Corresponding author: Xuefei Zhe (e-mail: xfzhe2-c@my.cityu.edu.hk).}}

\IEEEtitleabstractindextext{%
\begin{abstract}
Deep distance metric learning (DDML), which is proposed to learn image similarity metrics in an end-to-end manner based on the convolution neural network, has achieved encouraging results in many computer vision tasks. 
$L2$-normalization in the embedding space has been used to improve the performance of several DDML methods. 
However, the commonly used Euclidean distance is no longer an accurate metric for $L2$-normalized embedding space, i.e., a hyper-sphere. 
Another challenge of current DDML methods is that their loss functions are usually based on rigid data formats, such as the triplet tuple.
Thus, an extra process is needed to prepare data in specific formats.
In addition, their losses are obtained from a limited number of samples, which leads to a lack of the global view of the embedding space.
In this paper, we replace the Euclidean distance with the cosine similarity to better utilize the $L2$-normalization, which is able to attenuate the curse of dimensionality. 
More specifically, a novel loss function based on the von Mises-Fisher distribution is proposed to learn a compact hyper-spherical embedding space.
Moreover, a new efficient learning algorithm is developed to better capture the global structure of the embedding space.  
Experiments for both classification and retrieval tasks on several standard datasets show that our method achieves state-of-the-art performance with a simpler training procedure.  
Furthermore, we demonstrate that, even with a small number of convolutional layers, our model can still obtain significantly better classification performance than the widely used softmax loss.
\end{abstract}

}

\maketitle

\IEEEdisplaynontitleabstractindextext

%
\IEEEpeerreviewmaketitle

\section{Introduction}
\label{intro}
By combining deep learning with classical distance metric learning, deep metric learning achieves exciting results on many visual tasks. 
For example, by introducing the triplet loss to the deep learning framework, deep metric learning is found to be effective in face verification~\cite{schroff2015facenet,pr_face_ver,tip_face_discriminative}, person re-identification~\cite{personReID,pr_p_rid}, 3D object retrieval \cite{tip_3D} and image retrieval~\cite{zhao2015DSRH}. 

This new combination is also known as deep distance metric learning (DDML) or deep metric learning (DML). 
Among many DDML methods, the triplet embedding is the most widely used one.
For instance, deep metric learning with triplet shows competitive results on fine-grained visual categorization (FGVC) tasks \cite{cui2016fine}. 
FaceNet \cite{schroff2015facenet} uses the triplet loss with $L2$-normalization to handle face related tasks better. 
The work in~\cite{personReID} achieves the state-of-the-art performance on the re-identification problem. 
Besides triplet based methods, many other deep metric learning approaches have been proposed in recent years, e.g., the quintuplet loss~\cite{qlet} and lifted structure loss~\cite{DMLlifted}.
Most existing deep metric learning methods are designed based on the Euclidean distance. 
Several recent studies \cite{schroff2015facenet,DMLfacility} use $L2$-norm to normalize the embedding space. 
Though $L2$ normalization as an effective method is widely used to deal with the curse of dimensionality, little attention is paid to the fact that the normalization process projects a $p$-dim Euclidean space to a high dimension sphere, $\mathbb{S}^{p-1}$. 
In such a manifold, the Euclidean distance is no longer an accurate measurement. 
Therefore, it is reasonable to exploit geometric properties of the manifold, the hyper sphere here, for applying machine learning models to data with the unit norm \cite{sra2016directional}. 

Although there is little attention paid to the conflict between the Euclidean metric and the spherical embedding space in DDML, the studies on the data with the unit norm have achieved promising results in several disciplines \cite{banerjee2005DireClustering}, such as image clustering, text mining and gene expression analysis. 
This kind of data is also known as directional data for which the ``direction'' of data contains richer information than the ``magnitude''. 
It has already been shown by many studies that direction can represent data better than magnitude \cite{nguyen2010cosine,ChooseOrCompress}.

Besides the metric problem, it is widely complained that training with deep metric learning models is usually more complicated compared with the softmax loss.
Most of existing DDML methods rely on specific mini-batch formats, such as triplets \cite{schroff2015facenet} and n-pair tuples \cite{N-pairs}. 
Preparing these formats is very time-consuming.
Besides, the pair selection strategies have a significant influence on the final performance.
Other methods such as the algorithm in \cite{DMLfacility} do not require preparing data in any formats. However, extra steps are needed to solve their local facility function.

To address the above issues, we first replace the Euclidean distance in deep metric learn with the cosine similarity which is more suitable for $L2$-normalized embedding space. 
Then, by introducing the directional distribution, a novel deep metric learning model is proposed. 
More specifically, the von Mises-Fisher distribution, which can be treated as the Gaussian distribution for spherical data, is used to defined a new loss function named von Mises-Fishes loss (vMF loss) for our model. 
Besides the loss function, an alternative learning algorithm is proposed to efficiently train our model. 
Extensive experiments on both classification and retrieval tasks show that our method achieves the state-of-the-art performance with an simple training process. 
More over, our method obtains a better performance with shallow convolutional neural networks, which indicates our method has a wide potential use for many mobile applications.  
The main contributions of this paper are summarized as follows:
\begin{itemize}
\item To our knowledge, it is the first time that directional statistics is introduced to deep distance metric learning.
\item A novel loss function based on the von Mises-Fisher distribution is proposed for deep metric learning to learn an embedded probability space on a hyper sphere.
\item An alternative learning algorithm is proposed to train our model efficiently.
\end{itemize}

The rest of this paper is organized as follows. 
We firstly review some deep metric learning methods and briefly introduce directional statistics in machine learning in Sec. \ref{sec:review}.
Some preliminary knowledge about directional statistics is given in Sec. \ref{sec:preliminary}.
In Sec. \ref{sec:mehtod}, we present our deep metric learning model, followed by the learning algorithm. 
A toy example is used to show the embedding spaces at the end of this section.
Experiments on both classification and retrieve tasks are conducted in Sec. \ref{sec:exp}.
In Sec. \ref{sec:discussion}, we show the performance of vMF on convolutional neural networks with different depths and clustering performances with different clustering methods.
The conclusion and future works of this paper are drawn in the final section.

\section{Related Work}
\label{sec:review}
This section briefly discusses two active research areas highly related to our approach. 
After the review of the latest deep metric learning research, the directional statistics in machine learning will be briefly discussed.
\subsection{Deep Metric Learning}
Deep metric learning aims to learn a non-linear projection function which can transform an image from the pixel level to a discriminate space where samples from the same class will be gathered together, and samples from different classes will be pushed apart.
Recent studies of deep metric learning advance performance of many visual tasks, such as fine-grained categorization \cite{cui2016fine}, image retrieval \cite{DMLlifted}, deep hashing ~\cite{TIP_hashing}, face verification \cite{schroff2015facenet}, and person re-identification \cite{personReID,tip_reid}. 
In following parts, several latest deep metric learning methods are briefly reviewed.
\subsubsection{Triplet loss}
The main idea of the triplet loss is that distances between dissimilar pairs should be larger than distances between similar pairs with a margin $m$.
Given an anchor image denoted with $X_a$, a similar image $X_p$ and a dissimilar image $X_n$ are selected to form a triplet, the triplet loss function can be defined as:
\begin{equation}
\label{eq:triplet}
     \mathscr{L}=  [({d}(f_{\theta}(X_a),f_{\theta}(X_p))-{d}(f_{\theta}(X_a),f_{\theta}(X_n)) + m~]_+,
\end{equation}
where $d$ is a distance function and $f$ is a mapping function with parameters $\theta$. 
The $[\cdot]_+$ operator denotes the hinge function which equals to  $max(\cdot,0)$.
The performance of triplet-based deep metric learning highly relies on the quality of triplet pairs. 
Many methods have been proposed to deal with this, e.g., mining hard negative samples \cite{schroff2015facenet} or containing more pairs within a mini-batch \cite{QuintupletHuang2016,DMLlifted}. 
As a consequence, the training process becomes more complicated.
Moreover, the slow convergence problem of the triplet loss is widely complained. \cite{zhuang2016fast,rippel2015metric} 

\subsubsection{Lifted structured loss}
The lifted structured embedding \cite{DMLlifted} considers that each similar pairs should compare the distance with all the negative pairs.
The loss function is given as a $\log$-$sum$-$\exp$ formulation:
\begin{equation}
\begin{split}
\mathscr{L} = \frac{-1}{2|\mathcal{P}|} \sum_{(i,j)\in \mathcal{P}} &\left[ \log  \left(  \sum_{(i,k)\in \mathcal{N}} \exp{\{ \alpha - D_{i,j}\}} + \right. \right.   \\ 
& \left. \left. \sum_{(j,l)\in \mathcal{N}} \exp{\{ \alpha -D_{j,l}\}} \right) + D_{i,j} \right] ^2_+ ,   
\end{split}
\end{equation}
where $D_{i,j} = || f(X_i)-f(X_j)||_2$ is the Euclidean distance in the embedding space of samples $X_i$ and $X_j$,
 $\mathcal{P}$ is the set of positive (similar) pairs and $\mathcal{N}$ is the set of negative (dissimilar) pairs, and $\alpha$ is a constraint margin.
\subsubsection{N-pair loss}
N-pair \cite{N-pairs} considers to make full use of all pairs in a mini-batch. 
Given a $(N+1)$-pairs tuple: $\{{(X_i,X_i^+)} \}_{i=1}^N$, where samples from $N$ classes are selected and $X^i_+$ is a similar sample to $X_i$, the loss function can be formulated as follows:
\begin{equation}
\begin{split}
\mathscr{L} = & \frac{1}{N}  \sum_{i=1}^N \log \left( 1+ \sum\nolimits_{i\neq j}  \exp \left( f(X_i)^T f(X_j^+) \right. \right. - \\ 
& \left.  \left. f(X_i)^Tf(X_i^+)  \right) \right)  + \frac{\eta}{m} \sum_{i=1}^m||f(X_i)||_2,   
\end{split}
   \end{equation}
where $\eta$ is a $L2$ regularization on the embedding vectors.

\subsubsection{Magnet}
The Magnet \cite{rippel2015metric} suggests to punish overlaps between different clusters. The mini-batch structure is based on clusters. 
One cluster is first randomly selected, then $M-1$ closest clusters are chosen based on the distances on the embedding space. 
$D$ images per selected cluster will be randomly selected ($M\times D$ in total) to form a mini-batch. 
The loss function can be presented as 
\begin{equation}
    \mathscr{L}=\frac{1}{MD}\sum^{M}_{m=1}\sum^{D}_{d=1}\nonumber
\left\{\log\frac{\exp\{-\frac{1}{2\sigma^2}\|r_d^m -\hat{\mu}_m\|^2_2\}}{\sum\limits_{\hat{\mu}:C(\hat{\mu})\neq C(r_d^m)}\exp\{-\frac{1}{2\sigma^2}\|r_d^m -\hat{\mu}\|^2_2\}}\right\},
\end{equation}
where $r_d^m = f(X_d)$ is a point in the embedding space belong to $m$-th cluster.
The cluster center, $\hat{\mu}_m$, is estimated based on each batch data as follows:
\[
\hat{\mu}_m=\frac{1}{D}\sum_{d=1}^D r_d^m
\]
Though the convergence is claimed to be faster than triplet based methods, it is still very time consuming for generating mini-batch by retrieving images from adjacent clusters. 

\subsection{Directional Statistics in Machine Learning}
Besides visual tasks, the directional statistics methods in machine learning have been successfully introduced to many disciplines, e.g., text mining \cite{straub2015dirichlet}, gene expression analysis \cite{banerjee2005DireClustering}, and bio-medical data analysis \cite{lashkari2010fMRIdiscovering}.
More recently, an SNE method based on von Mises-Fisher distribution is proposed to deal with the high dimensional spherical data visualization. 
More detail can be found in a review paper \cite{sra2016directional}.
\section{Preliminary}
\label{sec:preliminary}
In this section, we will provide a brief introduction to the directional statistics.
\subsection{Directional Data}
The directional data refer to data with unit norm $||\textbf{x}||=1$, corresponding to points on a hyper sphere in $\mathbb{R}^p$. 
Usually we can transform any kinds of data to directional data by $L2$-normalization.
\subsection{Von Mises-Fisher Distribution}
\begin{figure}[h]
\centering
 \includegraphics[width = 0.4\textwidth]{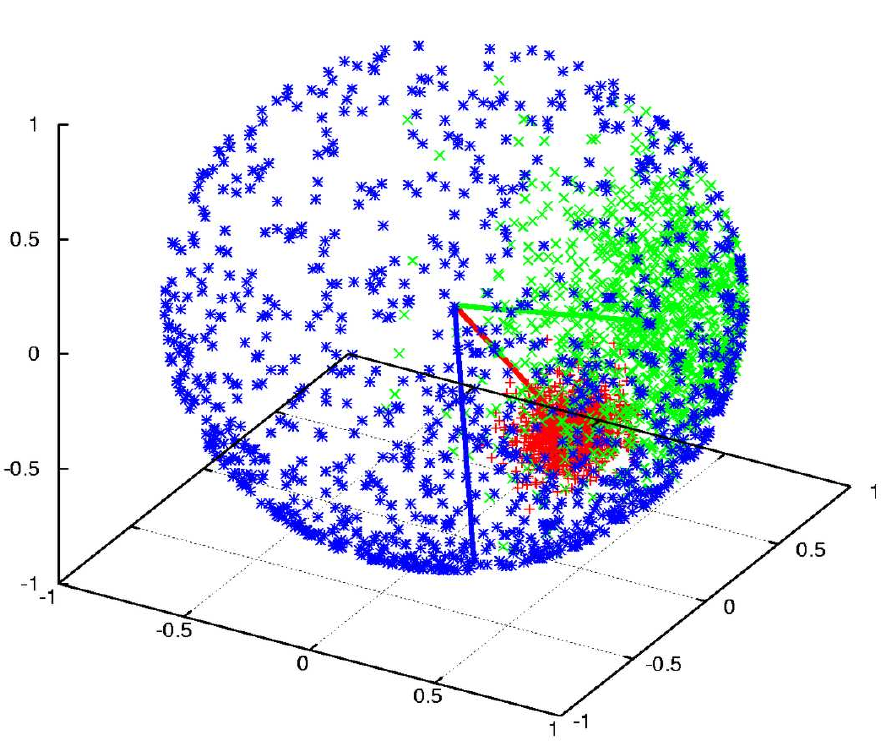}

   \caption{Points generated from three von Mises-Fisher distributions with the following parameters: {blue: $\kappa = 1$}, {green: $\kappa = 10$} and {red: $\kappa = 100$}. Better viewed in color.}
\label{fig:vmf-D}
\end{figure}

The von Mises-Fisher (vMF) distribution is a probability distribution in directional statistics for spherical data. 
The probability density function for a unit vector $\mathbf{r}$ in $\mathbb{S}^{p-1}$ is given as follows:
\begin{equation}
\mathit{f}_p(\mathbf{r};\bm{\mu},\kappa) = \mathit{Z}_p(\kappa)\exp(\kappa\bm{\mu}^T\mathbf{r}),
\end{equation}
where $\kappa\geq0$, $||\bm{\mu}||=1$ and the normalization parameter $\mathit{Z}_p(\kappa)$ is defined as follows:
\begin{equation}
\mathit{Z}_p(\kappa) = \frac{\kappa^{p/2-1}}{(2\pi)^{p/2}\mathit{I}_{p/2-1}(\kappa)},
\label{eq:C}
\end{equation}
where $\mathit{I}_v$ is the modified Bessel function of the first kind with order $v$. 
Similar to the Gaussian distribution, the vMF distribution is captured by two parameters: the mean direction $\bm{\mu}$ and the concentration parameter $\kappa$. Here $\kappa$ characterizes the tightness of the distribution around the mean direction $\bm{\mu}$. 
The larger value of ${\kappa}$ is, the more strongly the distribution is concentrated to the mean direction. 
Given $N$ sample points $\bm{r_i}$, the mean direction can be estimated as follows,
\begin{equation}
\bm{\mu} = \frac{\sum_{i=0}^N \mathbf{r}_i}{||\sum_{i=0}^N \mathbf{r}_i||}.
\end{equation}
A simple approximation estimate to $\bm{\kappa}$ is
\begin{equation}
\label{eq:kappa}
    \bm{\hat{\kappa}} = \frac{\bar{R}(p-\bar{R}^2)}{1-\bar{R}^2},
\end{equation}
where 
\[
    \bar{R}= \frac{\|\sum_{i=0}^N \bm{r_i} \|}{N}.
\]
Figure \ref{fig:vmf-D} shows points generated from different vMF distributions\footnote{This figure is adopted from the Wikipedia \url{https://en.wikipedia.org/wiki/Von_Mises-Fisher_distribution}.}. 
More information can be found in \textit{Directional Statistics}~\cite{directionalSta}.
\subsection{Hyper-surface Area of a Unit Hyper Sphere}
\begin{figure}[h]
\setlength{\abovecaptionskip}{0.cm}
\setlength{\belowcaptionskip}{-0.cm}
\centering
 \includegraphics[width=0.48\textwidth]{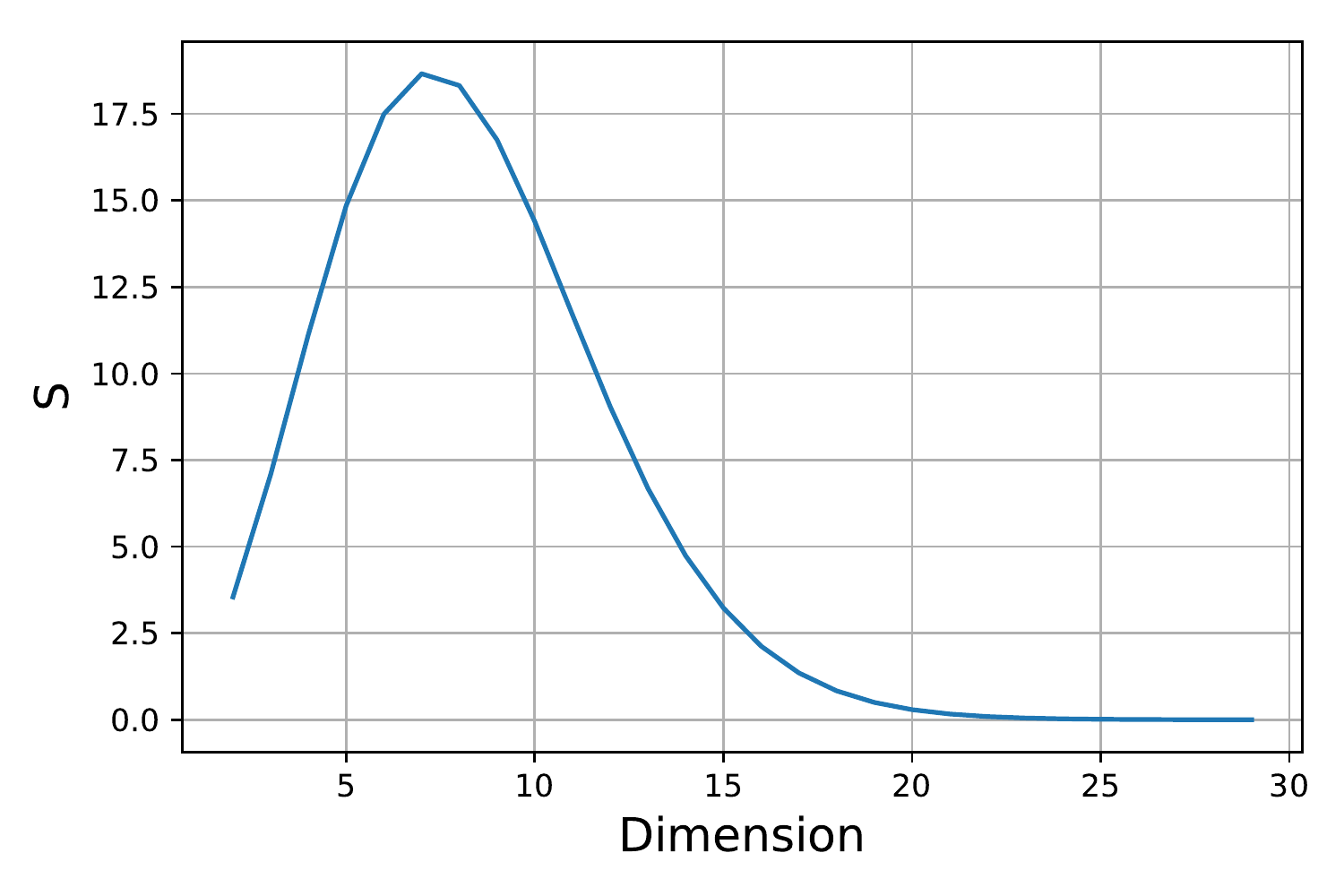}
 \caption{The area of a unit hyper sphere as a function of dimension.}
\label{fig:s_d}
\end{figure}

It is essential to understand the geometric properties of feature spaces for deep metric learning. 
However, for the feature space of the directional statistic, the property of the hyper-surface area of a unit hyper sphere is seldomly addressed. 
For a unit sphere in $\mathbb{R}^{p}$, the area of hyper-surface is given as follows~\cite{sphererBook},
\begin{equation}
S_p = \frac{2\pi^{p/2}}{\mathit{\Gamma(\frac{1}{2}p)}}=
\begin{cases}
\frac{2^{(p+1)/2}\pi^{(p-1)/2}}{(p-2)!!}& \quad \text{if } p \text{ is odd},\\
\frac{2^{p/2}\pi^{p/2}}{(\frac{1}{2}p-2)!}& \quad \text{if } p \text{ is even},
\end{cases}
\end{equation}
where $\mathit{\Gamma}$ refers to the Gamma function, $p!$ is the factorial and $p!!$ is the double factorial. 
We plot the area of hyper sphere in Figure~\ref{fig:s_d}. 
The area of a unit hyper sphere firstly increases to a maximum at dimension $p=7$ then decreases when $p$ increases. 
A distinct advantage of $L2$-normalization is that it can attenuate the curse of dimensionality. 
The volume of the Euclidean space increases exponentially as the dimension increase.
However, the input data are mapped to the surface of the surface of a hyper-sphere.
The area of the sphere increases as the sphere increases as the dimension increases initially, but then decreases as the dimensionality increases further.
Thus, a more compact feature space is created through the mapping.
\section{Method}
\label{sec:mehtod}
\begin{figure}[h]
\centering
 \includegraphics[width=0.8\linewidth]{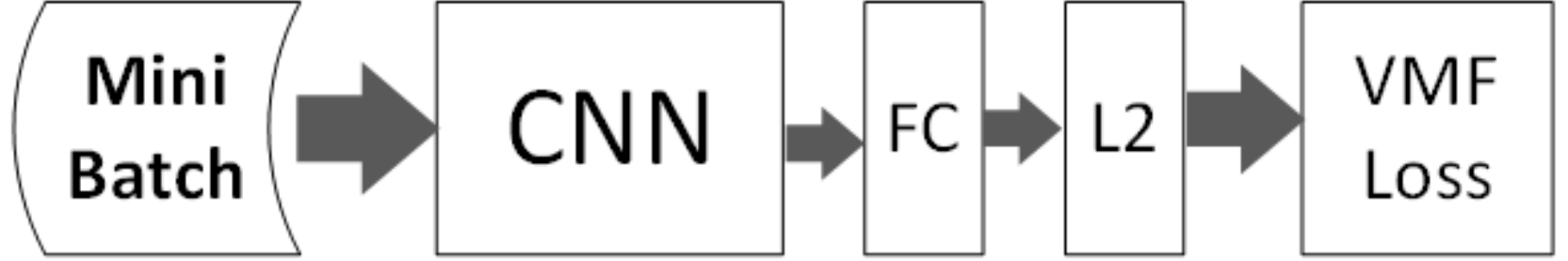}
   \caption{Overview of the model structure. A fully connected layer is  followed by an $L2$-normalization layer. The output feature will be used to compute the vMF loss.}
\label{fig:struct}
\end{figure}

In this section, we provide the problem definition of our deep metric learning model. 
Then we propose a loss function based on the von Mises-Fish (vMF) distribution along with an alternative learning algorithm. 
The overview of our model is presented in Figure~\ref{fig:struct}.
\subsection{Problem Definition}
Given a training set with $N$ pairs of labeled data $\{\mathbf{x}_n,y_n\}_{n=1}^N$ that belong to $C$ classes, our model aims to learn a non-linear mapping function $\mathit{f}(\cdot;\Theta)$, which is a convolutional neural network here. 
The learned non-linear mapping function projects an image ${\mathbf{x}}$ to a point, $\mathbf{r}$ and $\|\mathbf{r}\|=1$, on a $p$-dimension hyper sphere space dominated by $C$ learned vMF distributions $\{\kappa_i,\bm{\mu}_i\}_{i=1}^C$. 
In this probability space, a point $\mathbf{r}$ is assigned to class $c$ with the following normalized probability:
\begin{equation}
P(c|\mathbf{r},\{\kappa_i,\bm{\mu}_i\}_{i=1}^C) = \frac{\mathit{Z}_p(\kappa_c) \exp(\kappa_c \bm{\mu}_c^T\mathbf{r})}
{\sum_{i=1}^C\mathit{Z}_p(\kappa_i) \exp(\kappa_i \bm{\mu}_i^T\mathbf{r})},
\label{eq:vmf}
\end{equation}
where $\mathit{Z}_p(\kappa_i)$ is defined in Equation (\ref{eq:C}) and $||\mathbf{r}||=1$. 
The learned mapping function should project image data $\mathbf{x}_n$ to a point $\mathbf{r}_n$ in the new pace, which has the higher probability assigned to the right class $y_n=c$ than assigned to the other classes.
\begin{figure*}[ht]
\centering
 \includegraphics[width=0.8\linewidth]{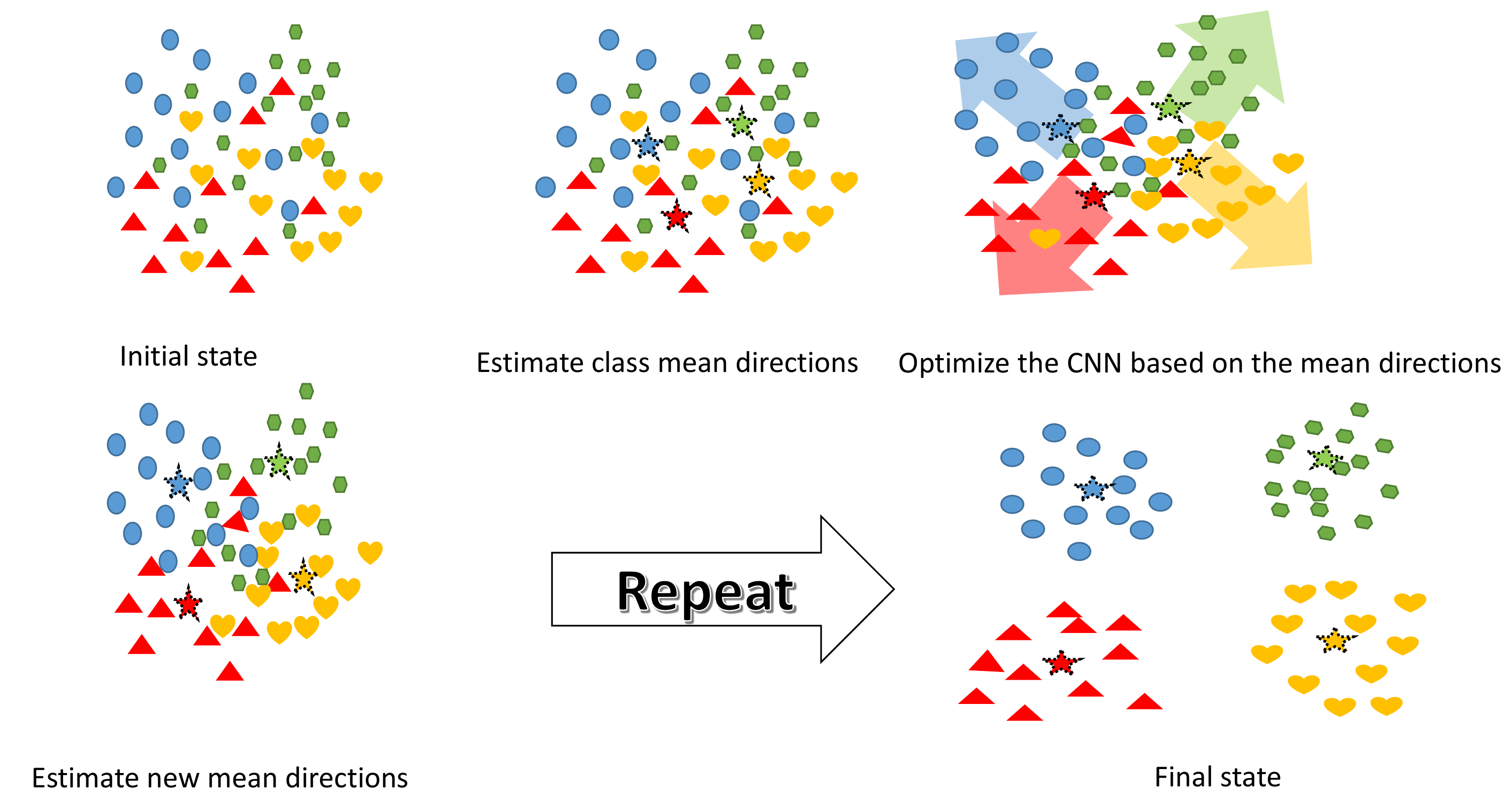}
   \caption{Illustration of the learning process. Different colors represent points from different classes. Symbol $\star$ represents one mean direction of a class.}
\label{fig:learning}
\end{figure*}
\subsection{Von Mises-Fisher Loss}
The above goal can be summarized as maximizing the following probability:
\begin{equation}
\begin{split}
\mathit{P}(Y|X;\Theta,\mathbf{U},K)& = \prod_{n=1}^N P(y_n|\mathbf{r}_n,\{\kappa_i,\bm{\mu}_i\}_{i=1}^C)\\
& = \prod_{n=1}^N \frac{\mathit{Z}_p(\kappa_{y_n}) \exp(\kappa_{y_n} \bm{\mu}_{y_n}^T\mathbf{r_n})}
{\sum_{i=1}^C\mathit{Z}_p(\kappa_i) \exp(\kappa_i \bm{\mu}_i^T\mathbf{r_n})},
\end{split}
\label{eq:Probability}
\end{equation}
where $X$ and $Y$ represent the training data and labels, $\Theta$ represents all the parameters in the mapping function $\mathit{f}$ and $\mathbf{U}=\{\bm{\mu}_i\}_{i=1}^C$, $K=\{\kappa_i\}_{i=1}^C$, $\mathbf{r}_n = \mathit{f}(\mathbf{x}_n;\Theta)$ and $||\mathbf{r}_n||=1, n=1,2,\dots,N$.
By taking the negative log-likelihood of Equation~(\ref{eq:Probability}), the objective function is obtained as follows:
\begin{equation}
\begin{aligned}
&\qquad~\underset{\Theta,\mathbf{U}}{\text{min}} & &\mathcal{J}= -\log \mathit{P}(Y|X;\Theta,\mathbf{U},K)\\
&&&=-\sum_{n=1}^N \log \frac{\mathit{Z}_p(\kappa_{y_n}) \exp(\kappa_{y_n} \bm{\mu}_{y_n}^T\mathbf{r_n})}
{\sum_{i=1}^M\mathit{Z}_p(\kappa_i) \exp(\kappa_i \bm{\mu}_i^T\mathbf{r_n})},\\
& \textrm{subject to} & &  ||\mathbf{r}_n||=1,\quad n=1,2,3,\dots,N \\
&  & & ||\bm{\mu}_i||=1,\quad i = 1,2,3,\dots,C,
\end{aligned}
\label{eq:objective}
\end{equation}
where $\mathbf{r}_n=\mathit{f}(\mathbf{x}_n;\theta)$, $\bm{r}_n,\bm{\mu}_i\in \mathbb{S}^{p-1}$, and the concentration parameters $\{\kappa_i\}_{i=1}^C$ is treated as hyper parameters to capture the divergence of each class. 
Here we simply set the same $\kappa$ for every class as a global scaling factor. Then Equation~(\ref{eq:vmf}) can be simplified to,
\begin{equation}
\mathscr{L} = -\log~\frac{ \exp(\kappa \bm{\mu}_{y_n}^T\mathbf{r_n})}
{\sum_{i=1}^C\exp(\kappa \bm{\mu}_i^T\mathbf{r_n})}.
\label{eq:l}
\end{equation}
We name this loss function as von Mises-Fisher Loss (vMF Loss) and it will be used for all experiments in this paper. 
Because Equations~(\ref{eq:objective}) and~(\ref{eq:l}) are differentiable, they can be used directly to train the neural network with the back propagation method.

\subsection{Learning}
Our vMF loss has two types of parameters: the mean directions of vMF distributions $\mathbf{U}$ and the parameters of the mapping function $\Theta$. 
It is difficult to optimize them simultaneously. 
So we apply an alternative training algorithm to learning these two types of parameters. 
In summary, we fix the mean directions of vMF distributions $\mathbf{U}$ when we train CNN by the mini-batch based stochastic gradient descent for some iterations. 
Then we use the learned CNN to forward pass all training data to obtain the representation vectors $\{\mathbf{r}_n\}$ and update the mean directions by the following expression,
\begin{equation}
{\bm{\mu}_i} = \frac{\sum_{n=1}^{N_i} \mathbf{r}_n^{(i)}}{||\sum_{n=1}^{N_i} \mathbf{r}_n^{(i)}||},
\label{eq:update}
\end{equation}
where $N_i$ is the number of samples that belongs to class $i$ and $\mathbf{r}_n^{(i)}=\mathit{f}(\mathbf{x}_n;\Theta)$. 
The overall learning algorithm is described in Algorithm\ref{algo:1} and illustrated in Figure \ref{fig:learning}.
\begin{algorithm}
\caption{ Learning algorithm}
\label{algo:1}
\begin{algorithmic}[1]
\STATE{\textbf{initialize} CNN parameters $\Theta$}
\REPEAT
  \STATE get $\mathbf{r}_n^{(m)}$ by forward passing training data 
    \STATE update the mean directions $\{\bm{\mu_i}\}$ by Eq. (\ref{eq:update})
    \STATE train CNN with Eq. (\ref{eq:l}) for $l$ iterations
\UNTIL{converge.}
\end{algorithmic}
\end{algorithm}

The training CNN step in the proposed alternative learning algorithm does not rely on any rigid data formats. 
Even we do not need to guarantee the class number in a mini-batch as in \cite{DMLfacility}. 
The training procedure is as simple as training with the softmax loss.
More important is that the global structure of the embedding space is represented by the estimated class mean directions.
For each sample, the loss computed with all mean directions is an approximation computed with the whole training set, which indicates that our method can capture the global information of the embedding space.
Previous works usually are insensible of the global landscape of the embedding space.
Taking the triplet loss as an example, the loss (see Equation (\ref{eq:triplet})) of an anchor image is only computed with one similar sample and one dissimilar sample.
The anchor image is kept in the dark about all other training points in the embedding space except the selected two.
That leads to a unstable training process, especially at the beginning, and the model is easy to be trapped in bad local optimal~\cite{rippel2015metric,DMLfacility}.
\subsection{Inference}
After training, we can predict the labels of samples by measuring the cosine similarity between sample features and the learned mean directions $\{\bm{u_i}\}$. 
Each Sample will be assigned with a class labels, of which the mean direction has the largest cosine similarity with the sample features. 
An interesting feature of this scheme is that, this prediction process also can enjoy the common nearest neighbor classifier based on the Euclidean distance. 
For two direction points ${\mathbf{r}_n}$ and ${\bm{\mu}_i}$, their distance is
\begin{equation}
d(\mathbf{r}_n,\bm{\mu}_i)=||\mathbf{r}_n - \bm{\mu}_i||_2^2 = ||\mathbf{r}_n||_2^2 + ||\bm{\mu}_i||_2^2-2\mathbf{r}_n^T \bm{\mu}_i.
\end{equation}
Considering $||\mathbf{r}_n||_2^2 = ||\bm{\mu}_i||_2^2=1$, their Euclidean distance reduces to $d(\mathbf{r}_n,\bm{\mu}_i)=2-2\cos<\mathbf{r}_n,\bm{\mu}_i>$, which means that their Euclidean distance is inversely proportional to the cosine similarity.
The largest cosine similarity pair can be found by looking for the pair with the shortest Euclidean distance.

\subsection{Visualization of the Learned Space}
\begin{figure*}[t]
\setlength{\abovecaptionskip}{0.cm}
\setlength{\belowcaptionskip}{-0.cm}
\begin{center}
\subfloat[Softmax] {\includegraphics[width=0.3\textwidth]{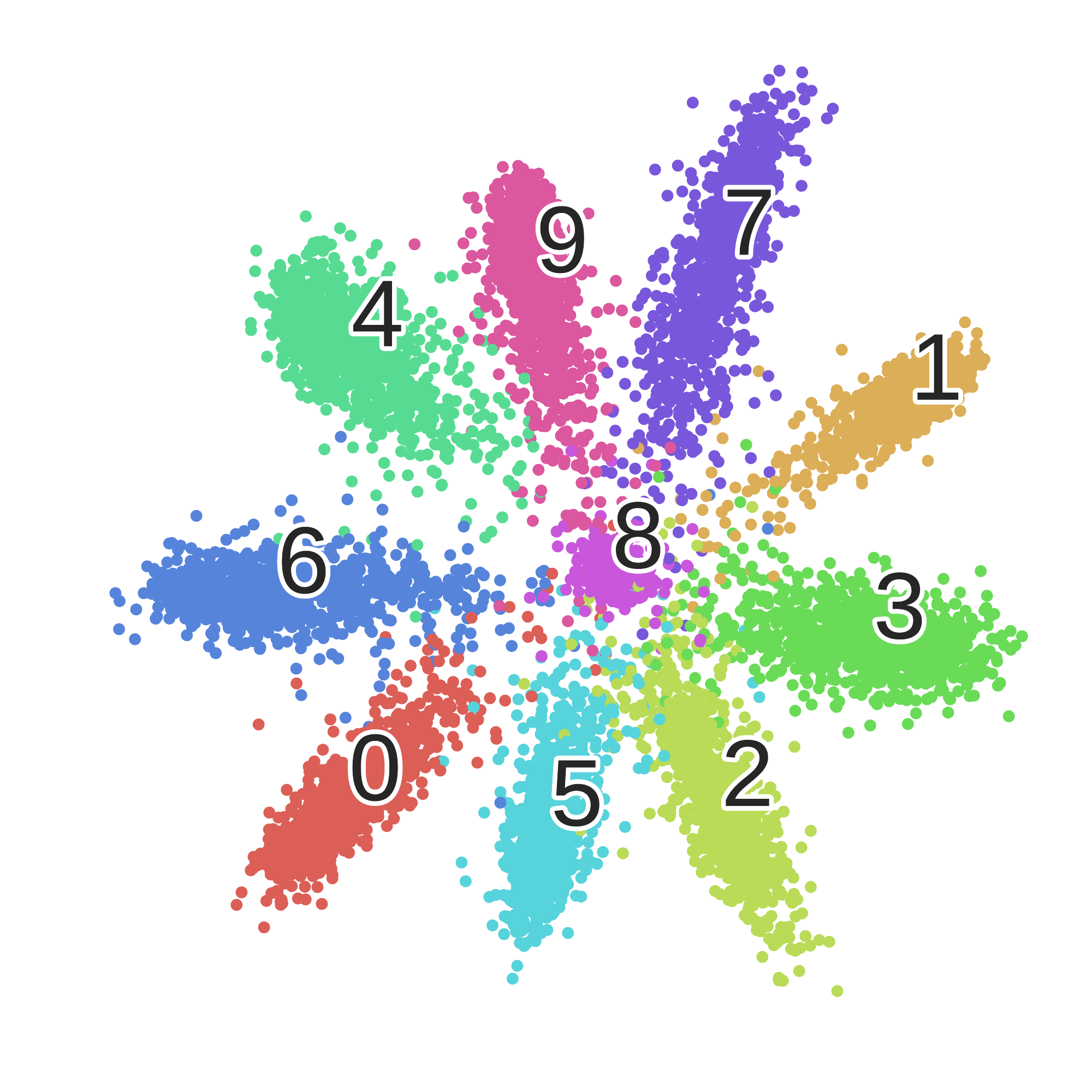}\label{fig:softmax}}
\subfloat[Triplet]{\includegraphics[width=0.3\textwidth]{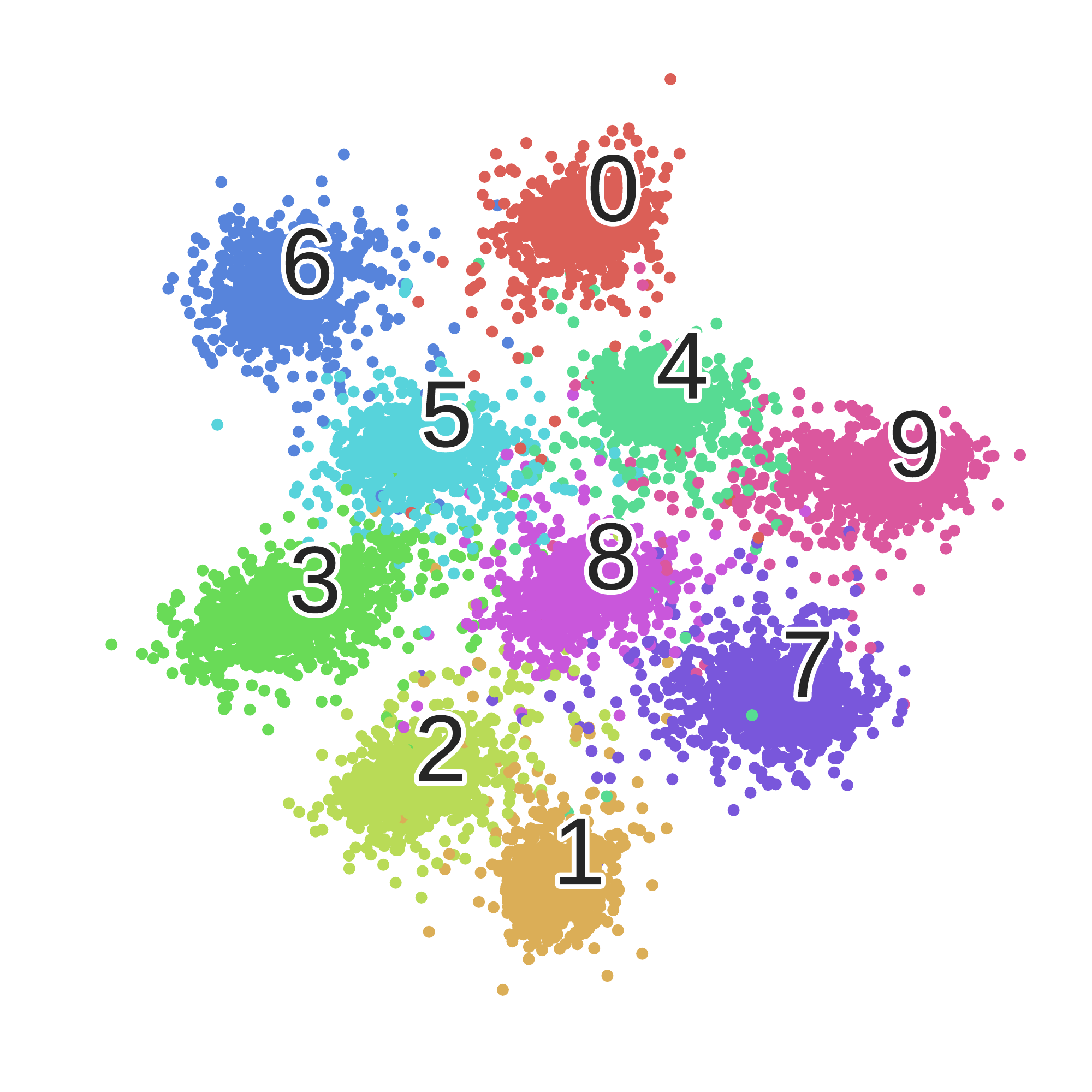}\label{fig:triplet}}
\subfloat[Center Loss with softmax]{\includegraphics[width=0.3\textwidth]{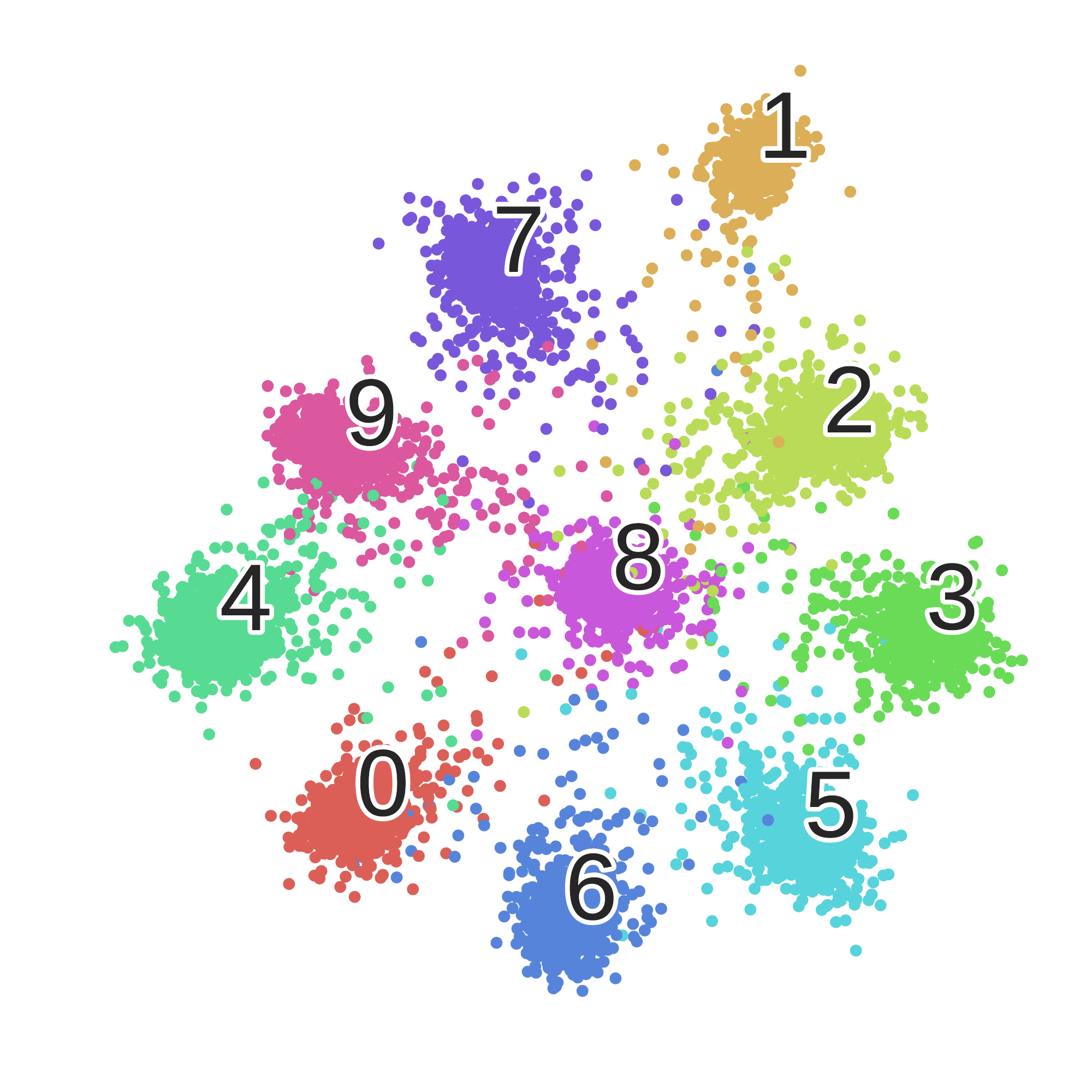}\label{fig:Centerl}}\\
\subfloat[vMF: 2D]{\includegraphics[width=0.25\textwidth]{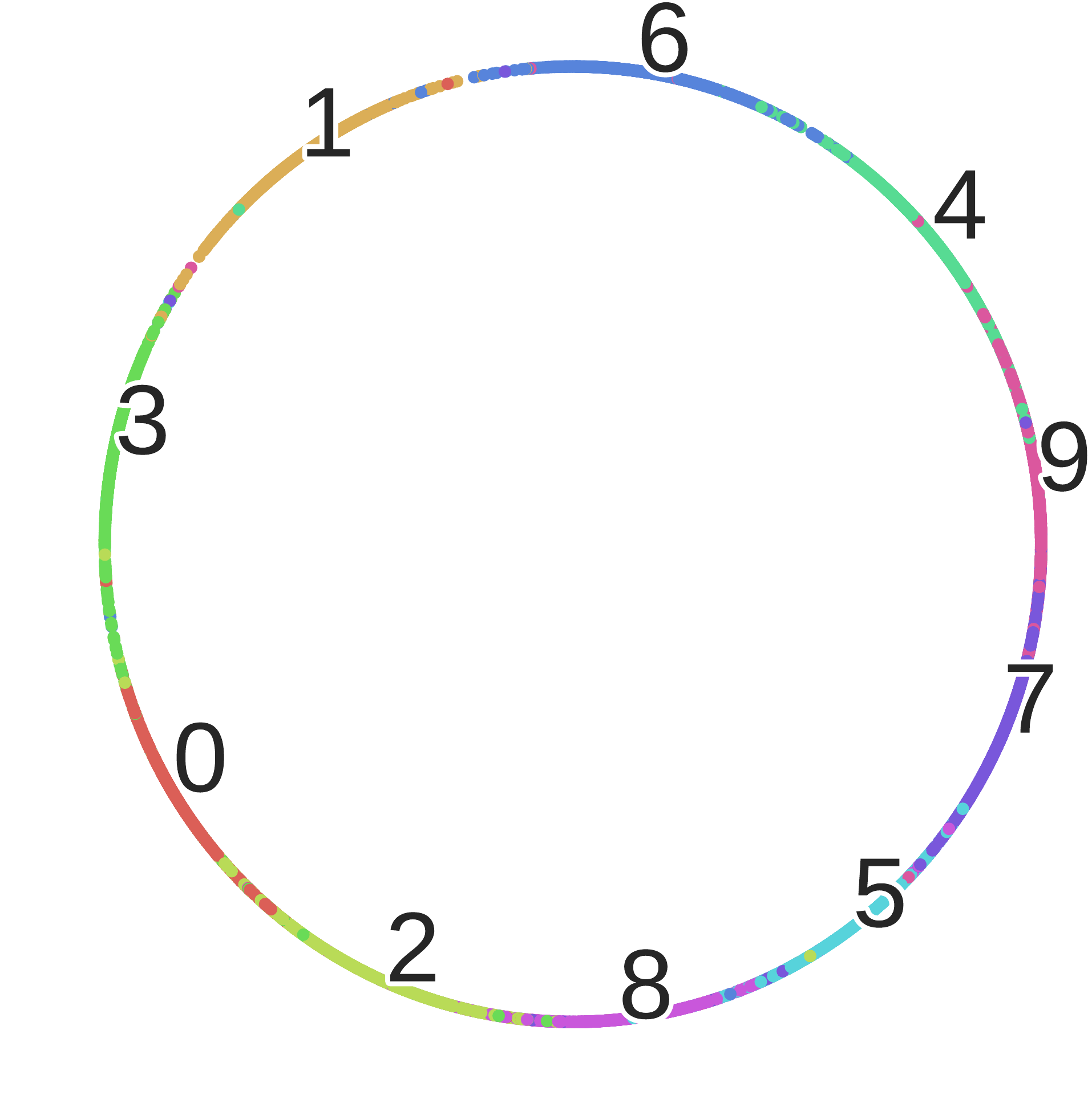}\label{fig:vmfLoss2d}}\quad
\subfloat[vMF: 3D, $\kappa = 15$]{\includegraphics[width=0.25\textwidth]{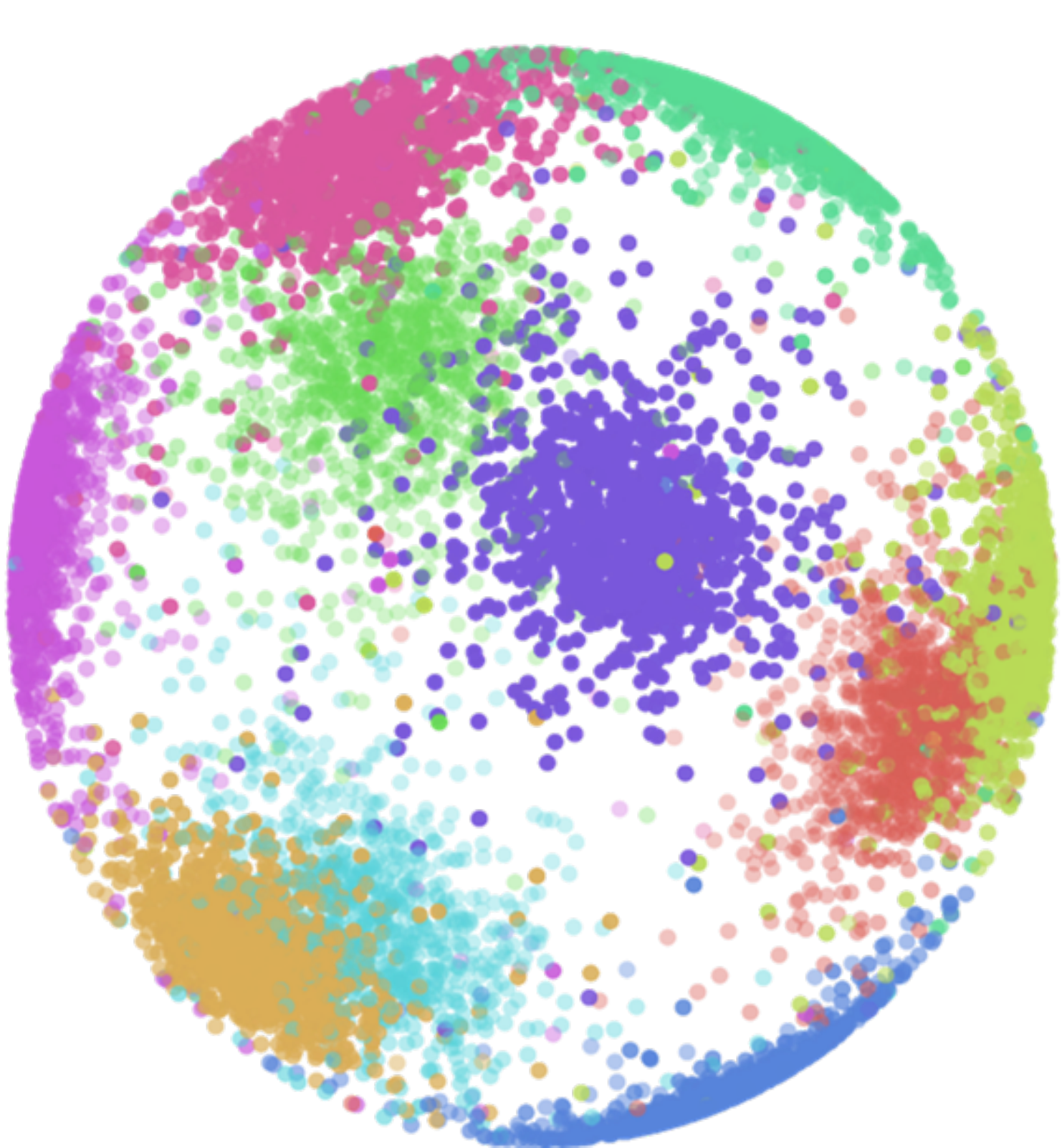}\label{fig:vmfLoss3d1}}\qquad
\subfloat[vMF: 3D, $\kappa = 70$]{\includegraphics[width=0.25\textwidth]{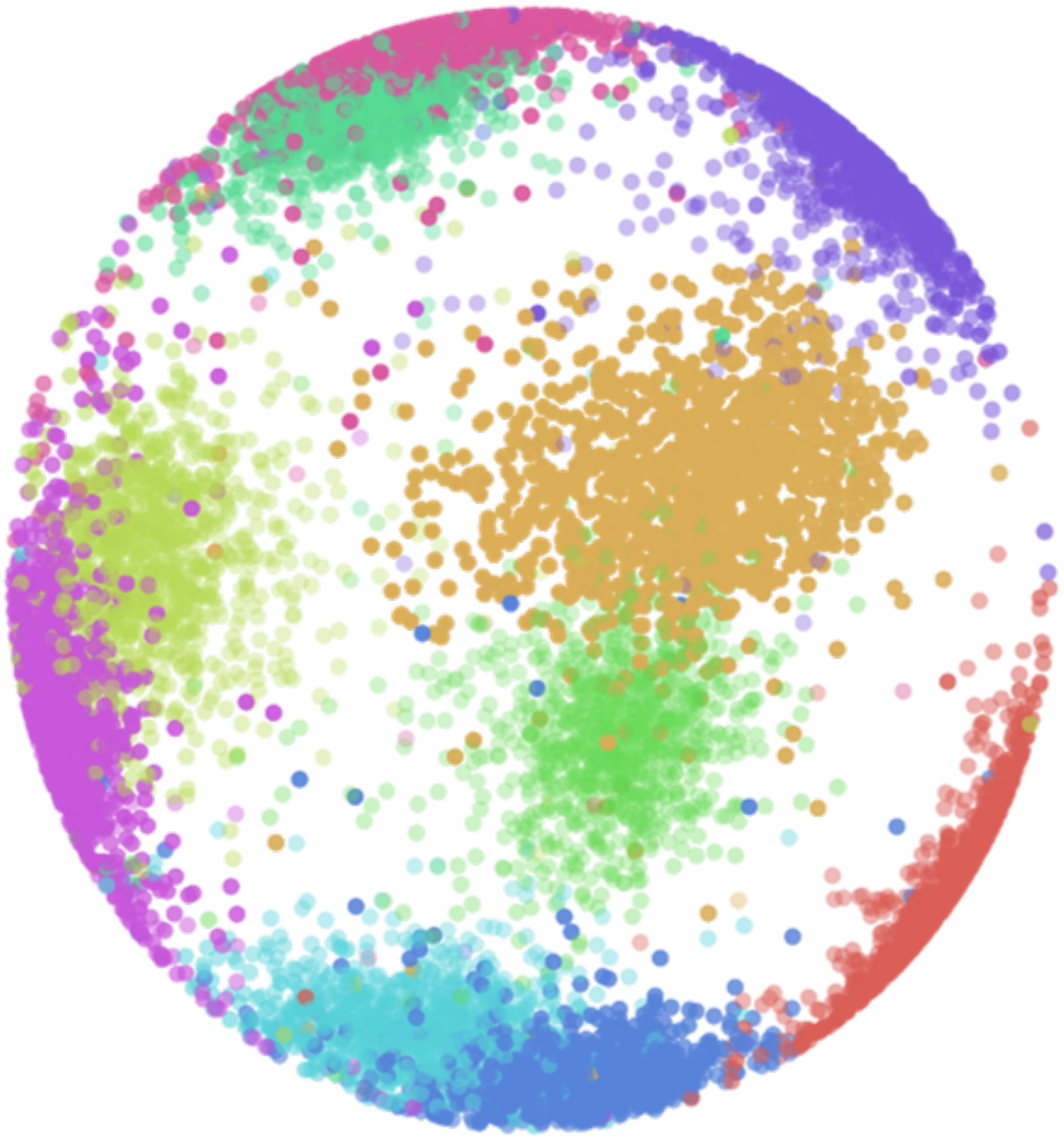}\label{fig:vmfLoss3d2}}\\
\caption{Visualization of embedding spaces with different loss functions. Better viewed in color.}
\end{center}

\label{fig:embedding}
\end{figure*}

We used the MNIST \cite{mnist} dataset as a toy example to give a simple comparison of different embedding spaces with different loss functions. 
A simple multi-layer perceptron (MLP)~\cite{mnist} is used here. 
The models are first trained with the training set then the learned model projects the validation set to the embedding spaces.
The embedding spaces of softmax, triplet loss~\cite{schroff2015facenet} and center loss~\cite{Center_lossWen2016} are also presented here for comparison.
From Figure \ref{fig:vmfLoss2d}, it can be seen that each class takes over an arc on a unit circle. 
Figures \ref{fig:vmfLoss3d1} and \ref{fig:vmfLoss3d2} provide the visualization results of different $\kappa$ in a $3D$ unit sphere. 
Both figures show that our model can learn a discriminant space for classification.  
%
\section{Experiments}
\label{sec:exp}
In this section, experiments on both classification and retrieval are conducted to verify the performance of our model for different tasks.
For the classification task, we test our model on three fine-grained data sets: Flower-102 \cite{flowers-102}, Oxford IIT Pet \cite{ox_pets} and Stanford Dog \cite{Dogs_FGVC2011}
For the retrieval task, the performance are evaluated base on three standard data sets: CUB-200-2011 \cite{bird200}, Cars196 \cite{cars196} and Stanford Online Products \cite{jegou2011product}. 
\subsection{Implementation Details}
We implement our model with MXNet\cite{chen2015mxnet}, an open source deep learning software. 
We initialize parameters of CNNs before the final fully connected layer with models pre-trained on the ImageNet \cite{imagenet_cvpr09}. 
The parameters in last fully-connected layer are initialize with Xavier initialization \cite{xavier}.
Experiments are run on a single NVIDIA GTX-1080 GPU.
\subsection{Classification on Fine-grained data}
Classifing object images in subordinate classes is known as fine-grained visual categorization (FGVC). 
For instance, in the general image classification task, a classifier only needs to recognize that it is a dog from a given picture of a dog.
But for the FGVC task, the classifier need to distinguish the breed which a dog belongs to, such as a Beagle or a Basset Hound. Many types of FGVC datasets have been built, including identification different species of animals and plants, classifying galaxies \cite{galaxy} and categorizing different air crafts \cite{AirCraft-grained}. 
One of main challenges for FGVC comes from the following two paradoxical properties. (1) FGVC has a high inter class similarity along with a large intra-class similarity. 
The difference caused by different sub classes may be smaller than that caused by various viewpoints or different shapes from the same sub-classes. 
(2) FGVC tasks usually have a large number of classes but a small number of samples for each class. Taking Oxford Flowers data set \cite{flowers-102} as an example, it contains 102 categories of flowers and only allows to use ten images for training and 10 for validation for each class. 
Compared with CIFAR-10 \cite{cifar10} ($6000$ images per class) and ImageNet \cite{imagenet_cvpr09} (around 1200 images for each class), the number of samples for each class in FGVC is challenging to learn a reliable model. 
Several deep metric learning approaches are proposed as promising solutions for FGVC tasks.

We evaluate our method on three FGVC datasets.  
To conduct a fair comparison, we follow the experiment setting in the Magnet \cite{rippel2015metric}.
The Inception with batch normalization \cite{ioffe2015batch} is used as our CNN part. 
The SGD with momentum is used to fine-tuning models. 
Rich data augmentation methods are used during training. 
The results of the Magnet and the triplet are directly cited from the original paper. 
The results of softmax and n-pairs are obtained by training with the same condition as the vMF loss. 
We set the dimension $p = 128$ and $\kappa = 15$ for three experiments. 
Neither the bounding box nor part information is used in our experiments.
 
\subsubsection{Flower-102}
The flower-102 data set contains $102$ categories of flowers. 
The number of images of each class varies from $40$ to $258$ and there are $8189$ images in total. 
For each category, there are $10$ images for training, $10$ images for validation, and the rest images for testing. 
We use all training and validation images as the training set. 
The concentration parameters $\{\bm{\mu}\}_{i=1}^{102}$ are updated every three epochs (around 100 iterations for 64 images per mini-batch ).  
The accuracy results are presented in Table \ref{tab:comflower}.
\begin{table}[h]
\centering
\caption{Classification results on Flower-102}
\label{tab:comflower}
\resizebox{0.35\textwidth}{!}{%
\begin{tabular}{|l|c|}
\hline
Mehtod                                & Mean Accuracy  \\ \hline
Softmax                               & 0.891          \\ \hline
Triplet                               & 0.830          \\ \hline
N-Piars                               & 0.885          \\ \hline
Magnet                                & 0.94           \\ \hline
VMF                                   & \textbf{0.956} \\ \hline \hline
Nilsback \cite{nilsback2008automated} & 0.856          \\ \hline
Sharif \cite{sharifflowers}           & 0.868          \\ \hline
Qian \cite{qian2015fine}              & 0.8945         \\ \hline
\end{tabular}%
}
\end{table}
\subsubsection{Oxford-IIIT Pet}
The Oxford-IIIT Pet \cite{ox_pets} data set provides image data from $37$ different breeds of dogs and cats, among which $25$ categories are dogs and 12 classes are cats. 
For each  breed, there are around $200$ images. 
Half number of images of each class are used for training and the rest are used for testing. 
Compared with previous flower-102 dataset, this pet data set has a larger variance in shapes, colors and scales. 
We update the mean direction of each class for every $100$ iterations.
\begin{table}[h]
\centering
\caption{Classification results on Oxford-IIIT Pet}
\label{tab:comPet}
\resizebox{0.35\textwidth}{!}{%
\begin{tabular}{|l|c|}
\hline
Method                            & Mean Accuracy  \\ \hline
Softmax                             & 0.879          \\ \hline
Triplet                             & 0.865          \\ \hline
N-pairs                             & 0.880          \\ \hline
Magnet                              & 0.894          \\ \hline
VMF                                 & \textbf{0.901} \\ \hline \hline
Mirray \cite{murray2014generalized} & 0.568          \\ \hline
Parkihi \cite{ox_pets}              & 0.592          \\ \hline
Qian \cite{qian2015fine}            & 0.812          \\ \hline
\end{tabular}%
}
\end{table}
\subsubsection{Stanford Dogs}
The Stanford Dogs data set \cite{Dogs_FGVC2011} contains $120$ categories of dog breads, and around $150$ images per class and $20,580$ in total.
Considering that this data set is a subset from ImageNet \cite{imagenet_cvpr09}, we follow the suggestion in the Magnet \cite{rippel2015metric} that a model that only trained two epochs on the ImageNet is used as initialization to avoid over-fitting.

The mean directions are updated after every epoch.
The results are reported in Table \ref{tab:comDogs}.
\begin{table}[h]
\centering
\caption{Classification results on Stanford Dogs}
\label{tab:comDogs}
\resizebox{0.35\textwidth}{!}{%
\begin{tabular}{|l|c|}
\hline
Method & Mean Accuracy \\ \hline
Softmax & 0.704 \\ \hline
Triplet & 0.642 \\ \hline
N-pairs & 0.688 \\ \hline
Magnet & 0.751 \\ \hline
VMF & \textbf{0.760} \\ \hline \hline
Qian \cite{qian2015fine} & 0.699 \\ \hline
Xie \cite{xie2015hyper} & 0.57 \\ \hline
Gavves \cite{gavves2015local} & 0.57 \\ \hline
\end{tabular}%
}
\end{table}
\subsubsection{Conclusion}
It can be found from above classification results that our vMF model achieves state-of-the-art performance on the tree fine-grained datasets. 
Only the Magnet and our method obtain better performance than the softmax for all three datasets.
On Flower-102, our model achieves $95.61\%$, which is $1.61\%$ higher than the result of the Magnet.
On Oxford-IIIT pet, our method has very close performance with the Magnet and is $0.7\%$ higher than the Magnet.
Our model exceeds the Magnet with $0.9\%$ on the Stanford Dogs.
\subsection{Retrieval Task}
\label{sec:retrieval_exp}
In this part, we focus on comparing retrieval performance with several state-of-the-art deep metric learning methods including (1) triplet learning with semi-hard negative mining strategy~\cite{schroff2015facenet}, (2) lifted structure embedding~\cite{DMLlifted}, (3) N-pairs metric loss~\cite{N-pairs} and (4) local facility clustering~\cite{DMLfacility}.
Because the Magnet only demonstrates the classification results in the original paper \cite{rippel2015metric}, we do not include the Magnet in this part.
The evaluation is conducted following the experiment protocol in~\cite{N-pairs,DMLlifted,DMLfacility}, which is $k$ nearest neighbor retrieval results on data whose classes are not appeared in the training sets.
The retrieval quality is measured by \textbf{Recall@K}~\cite{jegou2011product} (R@k), that is, the proportion of query images for which relevant items appear in the top $K$ neighbors. 
As suggested in~\cite{DMLfacility}, the embedding size does not significantly influence performance. 
The embedding size is fixed at $d=64$ for all experiments, which is the same as in \cite{DMLfacility}. 
Comparison results are based on three data sets: CUB200-2011~\cite{bird200}, Cars196~\cite{cars196} and Stanford Online Products~\cite{DMLlifted}.
All models are trained on data of first half classes then tested on images from the left half classes. 
Images are resized to $256\times256$ then cropped at $227\times227$. 
Random cropping and random mirroring are used for training data augmentation and single center crop are used for testing images.
The stochastic gradient descent (SGD) is used to training our vMF model for the retrieval task. 
The learning rate is reduced from $0.001$ to $0.00001$ with the batch size of $96$.
Due to the much larger number of classes in Stanford Online Product than in Cars196 and CUB-200-2011, the parameter $\kappa$ is experimentally set to $15$ for Stanford Online Product and $40$ for the other two datasets. 
The mean directions $\bm{\mu}$ are updated after every epoch.
Because the same experiment setting is applied as~\cite{DMLfacility}, the results of other deep metric learning methods are directly cited from \cite{DMLfacility}. 

\subsubsection{Cars196}
\begin{figure}[h]
\centering
 \includegraphics[width = 0.45\textwidth]{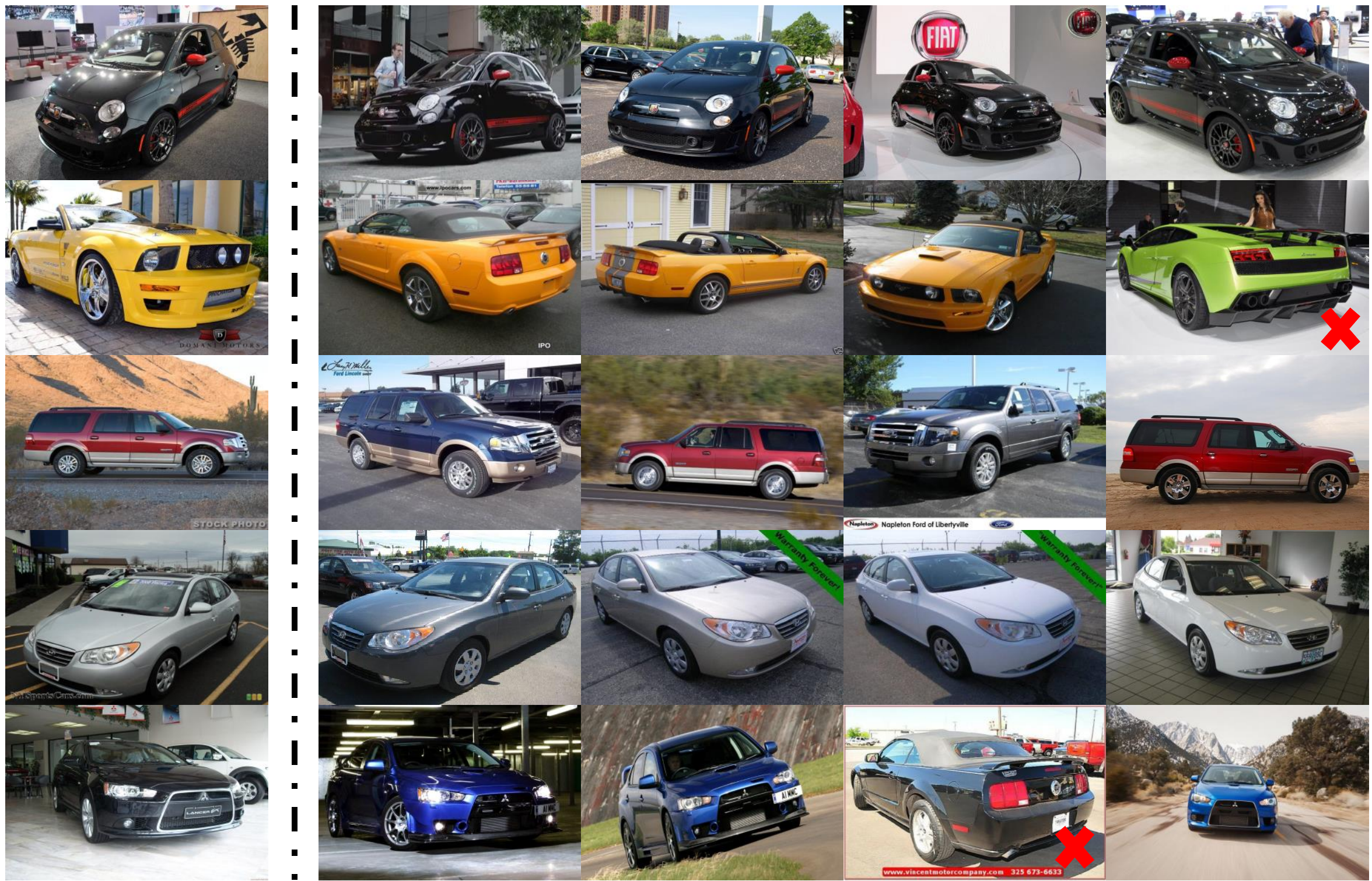}
   \caption{Successful retrieval examples of CARS196. The first image of every row is a query image. Images denoted with \textcolor{red}{x} are false items.}
\label{fig:car_success}
\end{figure}

\begin{figure}[h]
\centering
 \includegraphics[width = 0.5\textwidth]{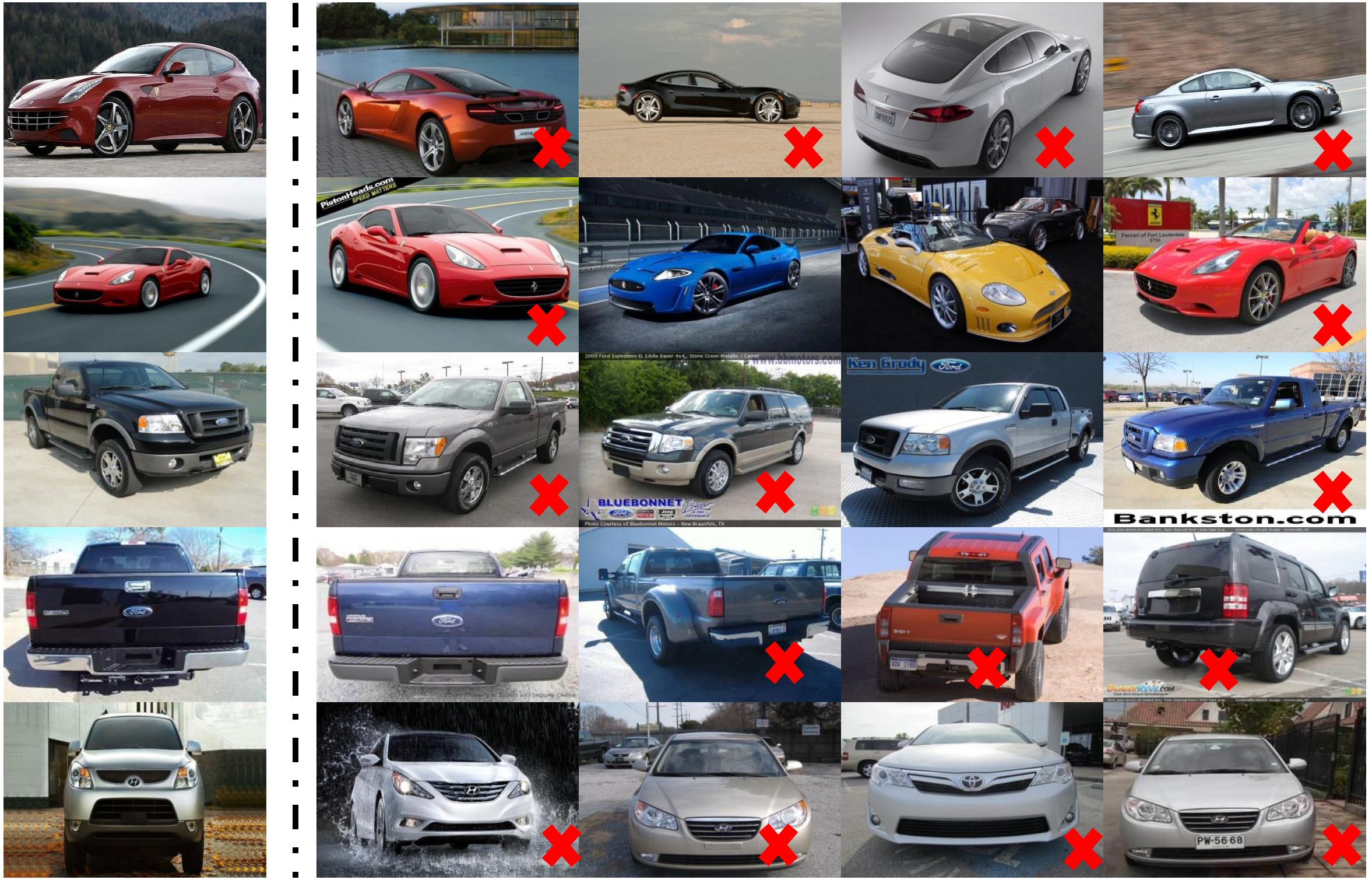}
   \caption{Failed retrieval examples of CARS196. The first image of every row is a query image. Images denoted with \textcolor{red}{x} are false items.}
\label{fig:car_failed}
\end{figure}

Cars196~\cite{cars196} is a large car dataset that includes $16,185$ images from $196$ classes of cars. 
Images from the first $98$ classes are used for training and the rest are used for testing. 
The results at $20,000$ iterations are presented in Table~\ref{tab:cars}. 
Our model outperforms other state-of-the-art methods.
The vMF achieves $62.34\%$ at R@1, which is $4\%$ higher than previous best one.
The successful and failed retrieval examples are presented in Figures~\ref{fig:car_success} and ~\ref{fig:car_failed}, respectively.
It can be observed that our model is sensitive to the view points to the cars.
\begin{table}[h]
\centering
\caption{Retrieval performance on Cars196 @$20k$ iterations.}
\label{tab:cars}
\resizebox{0.48\textwidth}{!}{%
\begin{tabular}{|c|c|c|c|c|}
\hline
Method & R@1 & R@2 & R@4 & R8 \\ \hline
Triplet semihard~\cite{schroff2015facenet} & 51.54 & 63.78 & 73.52 & 82.41 \\ \hline
Lifted structure~\cite{DMLlifted} & 52.98 & 65.70 & 76.01 & 84.27 \\ \hline
N-pairs~\cite{N-pairs} & 53.90 & 66.76 & 77.75 & 86.35 \\ \hline
Clustering~\cite{DMLfacility} & 58.11 & 70.64 & 80.27 & 87.81 \\ \hline
vMF & \textbf{62.34} & \textbf{73.39} & \textbf{82.52} & \textbf{89.45} \\ \hline
\end{tabular}%
}
\end{table}


\subsubsection{CUB-200-2011}
\begin{figure}[h]
\centering
 \includegraphics[width = 0.45\textwidth]{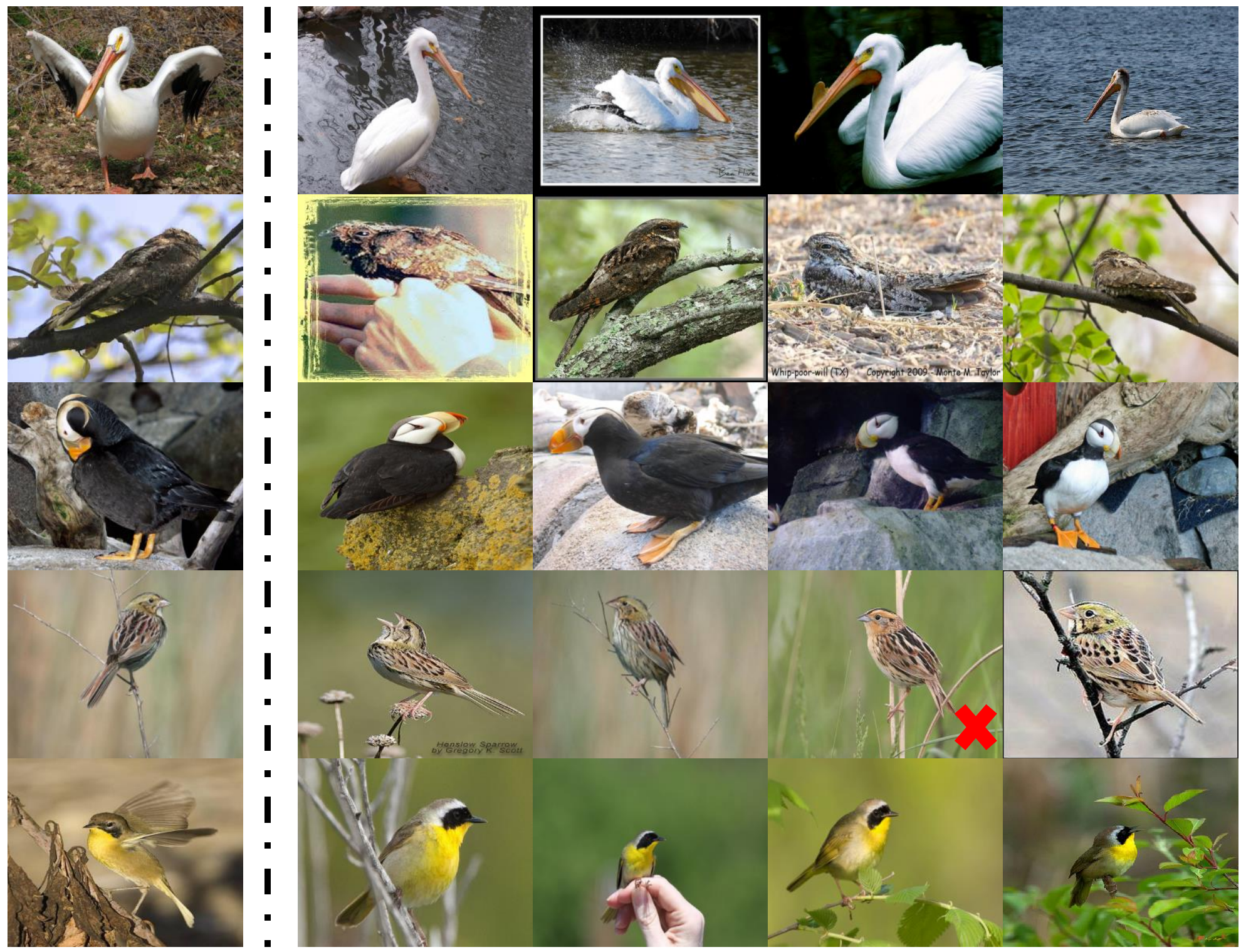}
   \caption{Successful retrieval examples of CUB-200-2011. The first image of every row is a query image. Images denoted with \textcolor{red}{x} are false items.}
\label{fig:birds_success}
\end{figure}

\begin{figure}[h]
\centering
 \includegraphics[width = 0.45\textwidth]{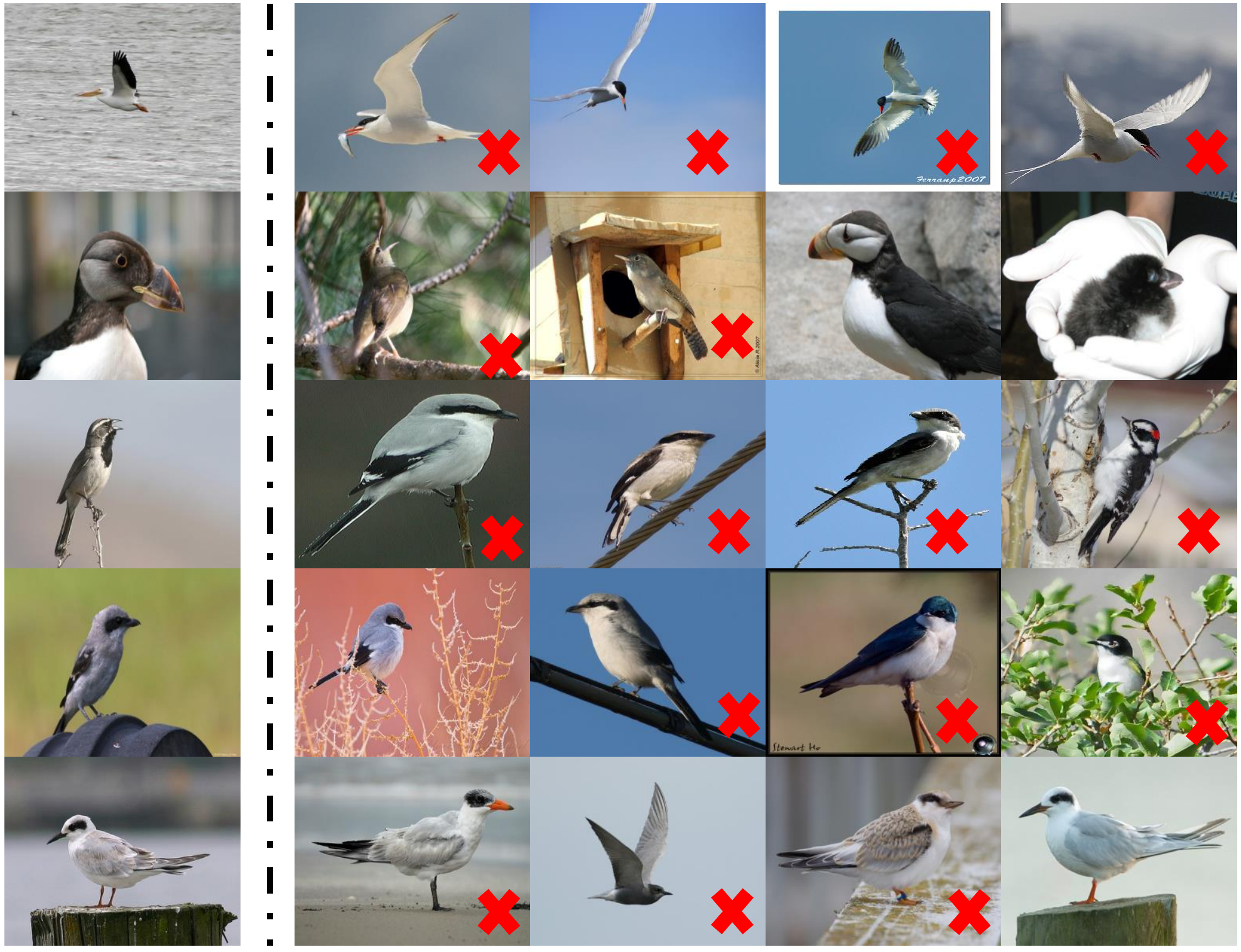}
   \caption{Failed retrieval examples of CUB-200-2011. The first image of every row is a query image. Images denoted with \textcolor{red}{x} are false items.}
\label{fig:birds_fails}
\end{figure}

CUB-200-2011~\cite{bird200} collects $11,718$ images from $200$ bird species.
From the dataset, $5,864$ images of the first $100$ categories are used for training and the rest $5,924$ images are used for testing. 
The quantitative result at $10,000$ iterations is shown in Table~\ref{tab:birds}.
It clearly shows that the vMF achieves the state-of-the-art performance.
Our method is $1.3\%$ higher than the Clustering method for R@1.
The successful and failed retrieval results are shown in Figures \ref{fig:birds_success} and \ref{fig:birds_fails}, respectively.
It can be found that large variants of different poses for the same class are one of the main challenges for CUB-200-2011.

\begin{table}[h]
\centering
\caption{Retrieval performance on CUB-200-2011 @$20k$ iterations.}
\label{tab:birds}
\resizebox{0.48\textwidth}{!}{%
\begin{tabular}{|c|c|c|c|c|}
\hline
Method                                     & R@1            & R@2            & R@4            & R8             \\ \hline
Triplet semihard~\cite{schroff2015facenet} & 42.59          & 55.03          & 66.44          & 77.23          \\ \hline
Lifted structure~\cite{DMLlifted}          & 43.57          & 56.55          & 68.59          & 79.63          \\ \hline
N-pairs~\cite{N-pairs}                     & 45.37          & 58.41          & 69.51          & 79.49          \\ \hline
Clustering~\cite{DMLfacility}              & 48.18          & 61.44          & 71.83          & 81.92          \\ \hline
vMF                                        & \textbf{49.48} & \textbf{61.77} & \textbf{73.35} & \textbf{83.17} \\ \hline
\end{tabular}%
}
\end{table}
\subsubsection{Stanford Online Products}
\begin{figure}[h]
\centering
 \includegraphics[width = 0.48\textwidth]{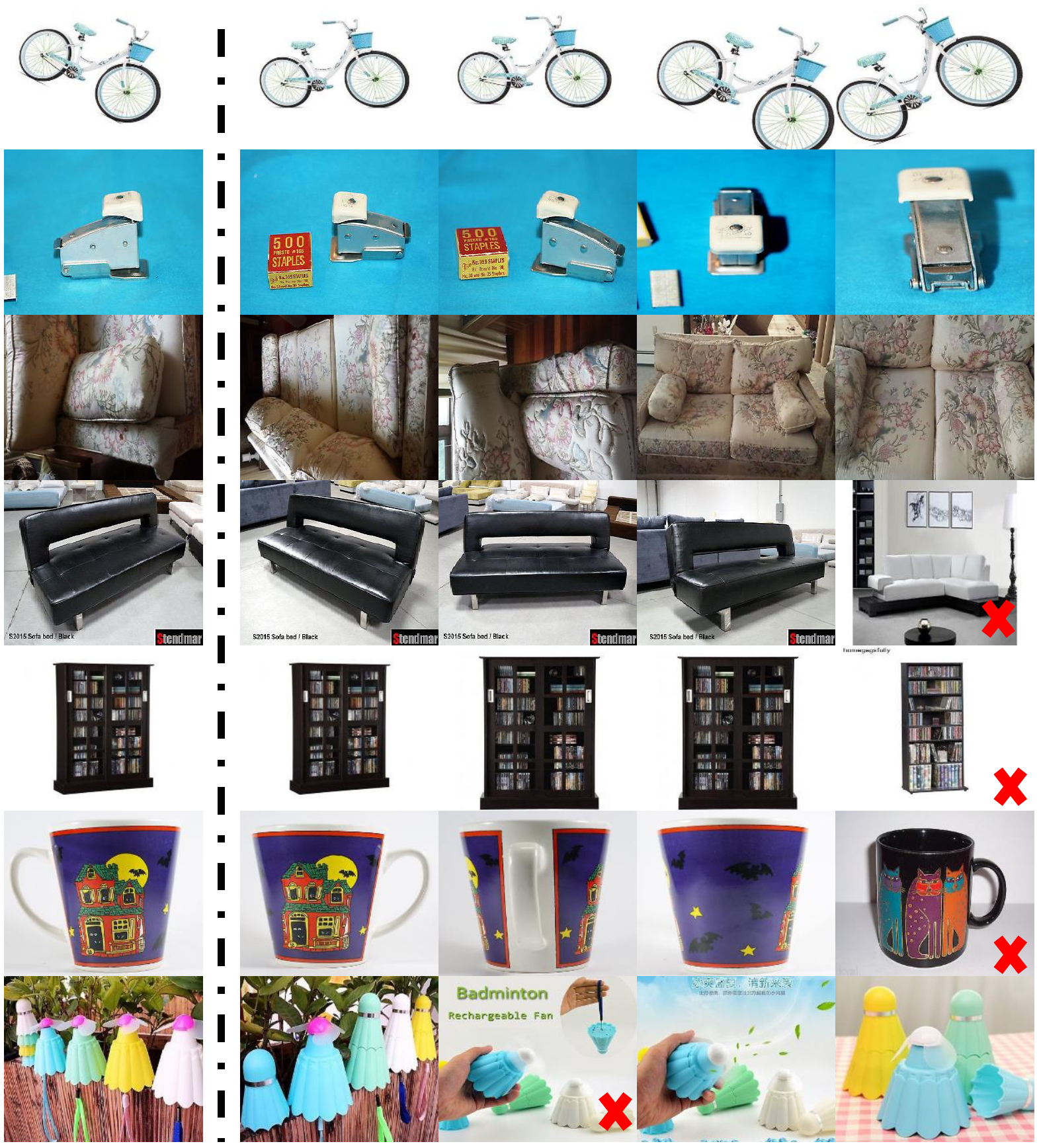}
   \caption{Successful retrieval examples of Stanford online products. The first image of every row is a query image. Images denoted with \textcolor{red}{x} are false items.}
\label{fig:products_success}
\end{figure}

\begin{figure}[h]
\centering
 \includegraphics[width = 0.48\textwidth]{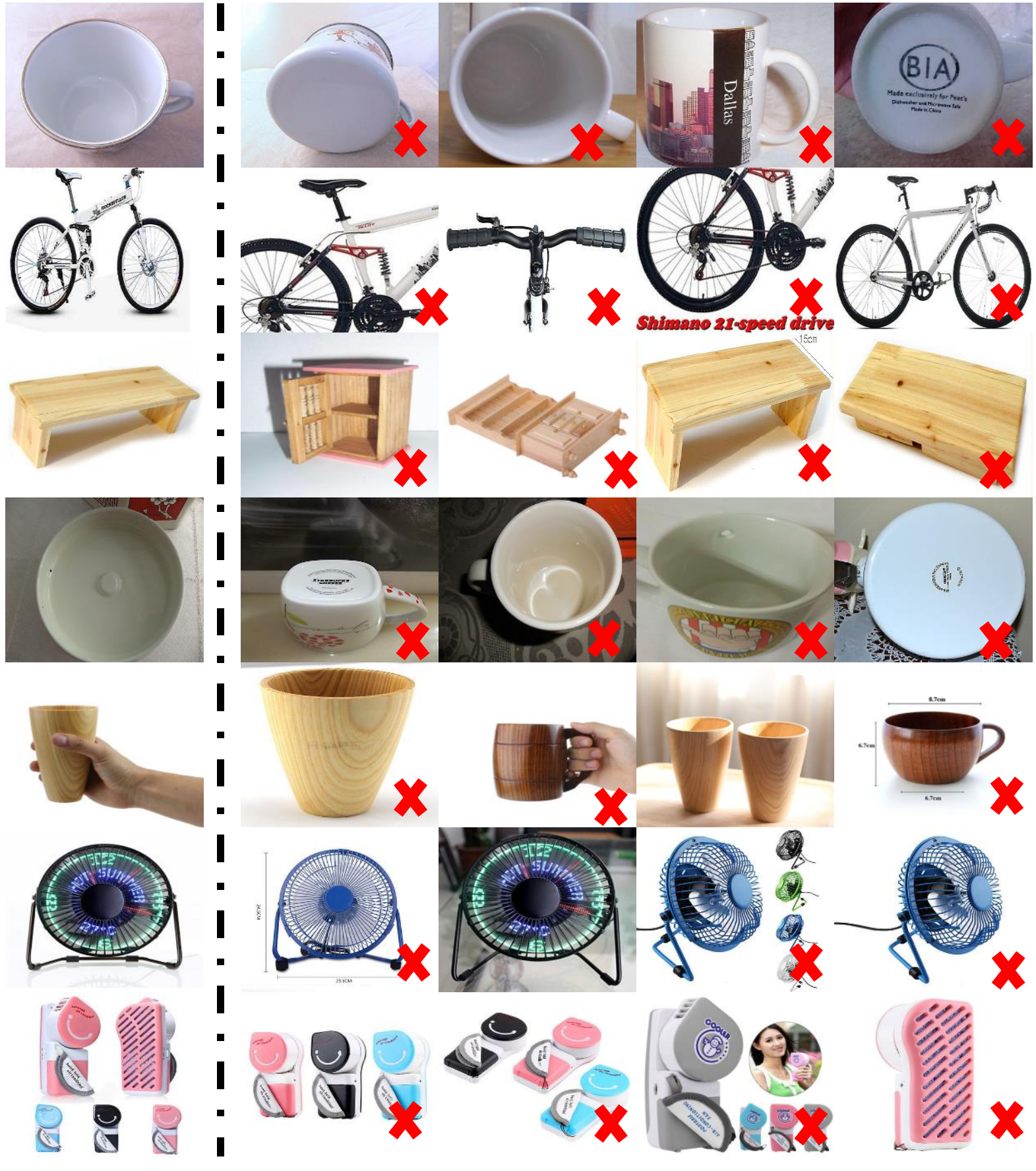}
   \caption{Failed retrieval examples of Stanford online products. The first image of every row is a query image. Images denoted with \textcolor{red}{x} are false items.}
\label{fig:products_bad}
\end{figure}

The Stanford online Products dataset~\cite{DMLlifted} includes $120,053$ images of $22,634$ online products collected from eBay.com. 
Each product is treated as one class.
The average number of images for each product is around $5.3$.
The images of the first $11,318$ products are used as training samples in this experiment.
The rest data are used for testing.
The $R@K$ metric results are presented in Table~\ref{tab:products}. 
Our vMF method surpasses other compared latest deep metric learning methods. 
Some successful and failed retrieval examples are presented in Figures \ref{fig:products_success} and \ref{fig:products_bad}, respectively.
Though the viewpoints and product poses change dramatically, our method still can return the correct images from the same class.
Most of false retrieval results come from the products from the same category but belong to different products with few differences.

\begin{table}[h]
\centering
\caption{Retrieval performance on Stanford Online Products @$20k$ iterations.}
\label{tab:products}
\resizebox{0.48\textwidth}{!}{%
\begin{tabular}{|c|c|c|c|}
\hline
Method                                     & R@1            & R@10           & R@100          \\ \hline
Triplet semihard~\cite{schroff2015facenet} & 66.67          & 82.39          & 91.85          \\\hline
Lifted structure~\cite{DMLlifted}          & 62.46          & 80.81          & 91.93          \\ \hline
N-pairs~\cite{N-pairs}                     & 66.41          & 83.24          & 93.00          \\ \hline
Clustering~\cite{DMLfacility}              & 67.02          & 83.65          & 93.23          \\\hline
vMF                                        & \textbf{67.53} & \textbf{84.51} & \textbf{93.36}\\\hline
\end{tabular}%
}
\end{table}

\section{Discussion}
\label{sec:discussion}
\subsection{Classification Performance on Different Depths}
\begin{figure}[H]
\centering
 \includegraphics[width = 0.45\textwidth]{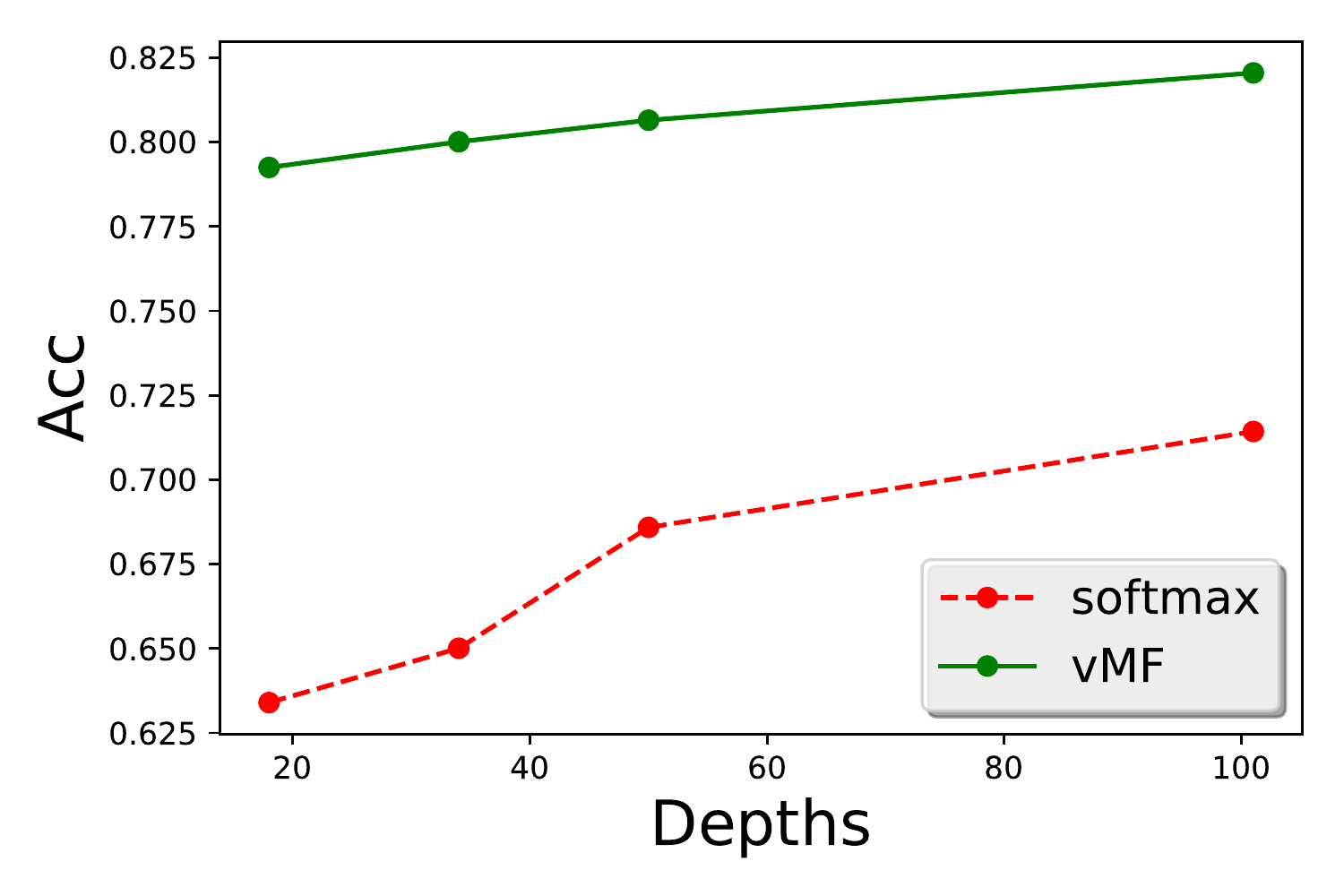}
  \caption{Mean Accuracy on CIFAR-100.}
\label{fig:depths}
\end{figure}

It has been a trend that improving the performance of CNN by making it deeper and more complicated.
However, it is very challenge to apply large networks to many real applications, such as mobile phones and robotics.
As an advantages of our model, vMF can achieves better results with ``shallow'' networks compared with wildly used softmax.
To demonstrate this property, we evaluate our model with different depths of ResNet \cite{resnet} on the CIFAR-100 dataset \cite{cifar10}. 
The CIFAR-100 provides $60,000$ images from $100$ classes ($600$ image per class). 
The train and test sets contain $50,000$ and $10,000$ images respectively.
We firstly resize images to $240\times240$, then the random cropping and random mirroring are used as data augment. 
The mean directions are updated after every epoch.
We report the quantitative results in Table \ref{tab:vmfVsSoftmax} and plot results in Figure \ref{fig:depths} for easy comparison.
It can be found that our method surpasses softmax with a significantly margin for all tested depths.
VMF is $15.8\%$ higher than softmax for ResNet-18 and $10.6\%$ higher for ResNet-101.
Moreover, vMF obtains $79.25\%$ accuracy for ResNet-18, which is $7.8\%$ higher than softmax with ResNet-101.
It clearly shows that our method has a higher depth efficiency.
\begin{table}[h]
\centering
\caption{Classification on different depths}
\label{tab:vmfVsSoftmax}
\resizebox{0.45\textwidth}{!}{%
\begin{tabular}{|c|c|c|c|}
\hline
\multicolumn{1}{|l|}{\multirow{2}{*}{Structure}} & \multirow{2}{*}{Depth} & \multicolumn{2}{c|}{Acc} \\ \cline{3-4} 
\multicolumn{1}{|l|}{}                           &                        & VMF        & Softmax     \\ \hline
\multirow{4}{*}{ResNet}                          & 18                     & 0.7925     & 0.6340      \\ \cline{2-4} 
                                                 & 34                     & 0.8001     & 0.6501      \\ \cline{2-4} 
                                                 & 50                     & 0.8065     & 0.6859      \\ \cline{2-4} 
                                                 & 101                    & 0.8205     & 0.7143      \\ \hline
\end{tabular}%
}
\end{table}

\subsection{Clustering Performance}
The clustering task is another important application for the deep distance metric learning. 
In practice, the clustering performance not only depends on the learned features, but also is influenced by how well the clustering methods works with the features.
In this parts, we show that, with the help of von Mises-Fisher distribution-based clustering methods, our method can outperform other state-of-the-art DDML for the clustering task.

In previous works, \cite{N-pairs,DMLlifted,DMLfacility}, all clustering performance is conducted with the affinity propagation clustering \cite{frey2007clustering} with bisection method\footnote{\url{http://www.psi.toronto.edu/affinitypropagation/apclusterK.m}} (apk).
However, it is worthwhile to choose the clustering methods that can better exploit the geometric properties of the embedding space.
Besides the apk method, we presents the clustering results with other three clustering methods based on the von Mises-Fisher distribution\footnote{\url{https://github.com/clara-labs/spherecluster}}  \cite{banerjee2005DireClustering}, including: (1) spherical k-means, (2) mixture of vMF-soft, (3) mixture of vMF-hard. 
The features of testing data are first extracted by models trained in Sec \ref{sec:retrieval_exp} then used for different clustering methods to obtain the cluster labels.  
The clustering performance is measured by normalized mutual information (NMI) \cite{DMLlifted}. 
We refer the clustering performance reported in \cite{DMLfacility} as a comparison, which is the previous best one.
The results are presented in Table \ref{tab:clustering}. 
It can be found that, using the apk method, our model achieves $0.5915$ and $0.5889$ for the Cars196 and CUB-200-2011, respectively.
These results are close to the state-of-the-art performance achieved by \cite{DMLfacility}.
More over, Working with all three von Mises-Fisher distribution-based clustering methods, our model can outperforms the clustering method \cite{DMLfacility} with apk.
Using movMF-hard clustering method, our method surpasses the clustering~\cite{DMLfacility} with $3.39\%$ and $5.10\%$ for the Cars196 and CUB-200-2011, respectively. 
\begin{table}[h]
\centering
\caption{Clustering performance}
\label{tab:clustering}
\resizebox{0.5\textwidth}{!}{%
\begin{tabular}{|c|c|c|c|}
\hline
DDML                        & Clustering methods & Cars196 & CUB-200-2011 \\ \hline
\multirow{4}{*}{vMF (ours)} & apk                & 59.15   & 58.89        \\ \cline{2-4} 
                            & spherical k-means  & 60.08   & 62.23        \\ \cline{2-4} 
                            & movMF-soft         & 61.20   & 62.20        \\ \cline{2-4} 
                            & movMF-hard         & 62.43   & 64.33        \\ \hline
Clustering~\cite{DMLfacility}                  & apk                & 59.04   & 59.23        \\ \hline
\end{tabular}%
}
\end{table}
\subsection{Impacts of $\kappa$}
\begin{figure}[h]
\centering
 \includegraphics[width = 0.43\textwidth]{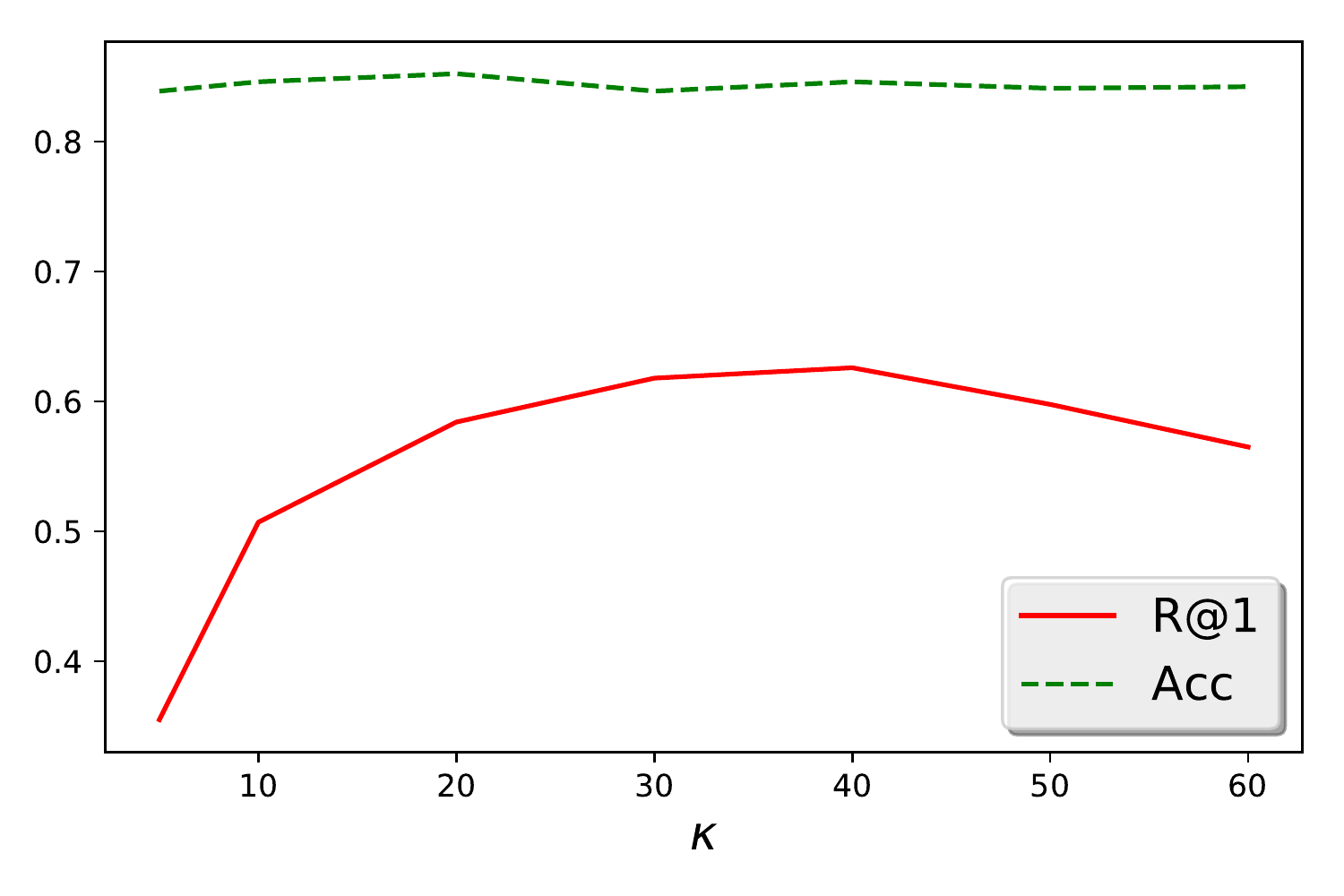}
\caption{Accuracy and Recall@1.}
\label{fig:acc_recall}
\end{figure}

In this part, we experimentally investigate how the hyper parameter $\kappa$ influences our model. 
The Cars196 \cite{cars196} is used as an exemplar dataset. 
To simultaneously monitor classification and retrieval performances, $20$ images per class of first $98$ classes are randomly selected as testing set for classification. 
The rest data in the first $98$ classes are used for training. 
All images in rest $98$ classes are used to test retrieval performance.
The mean accuracy and Recall@$1$ (R@$1$) are presented as measurements for classification and retrieval, respectively.
The results are plotted in Figure \ref{fig:acc_recall}.
It can be observed the value of $\kappa$ has little influence for classification performance.
However, the selection of $\kappa$ significantly impacts the retrieval performance. 
The retrieval performance first increases with the value of $\kappa$, and reaches a peak around $30\sim40$.
Then recall@$1$ goes down when $\kappa$ continues increasing.

\begin{figure}[h]
\centering
 \includegraphics[width = 0.49\textwidth]{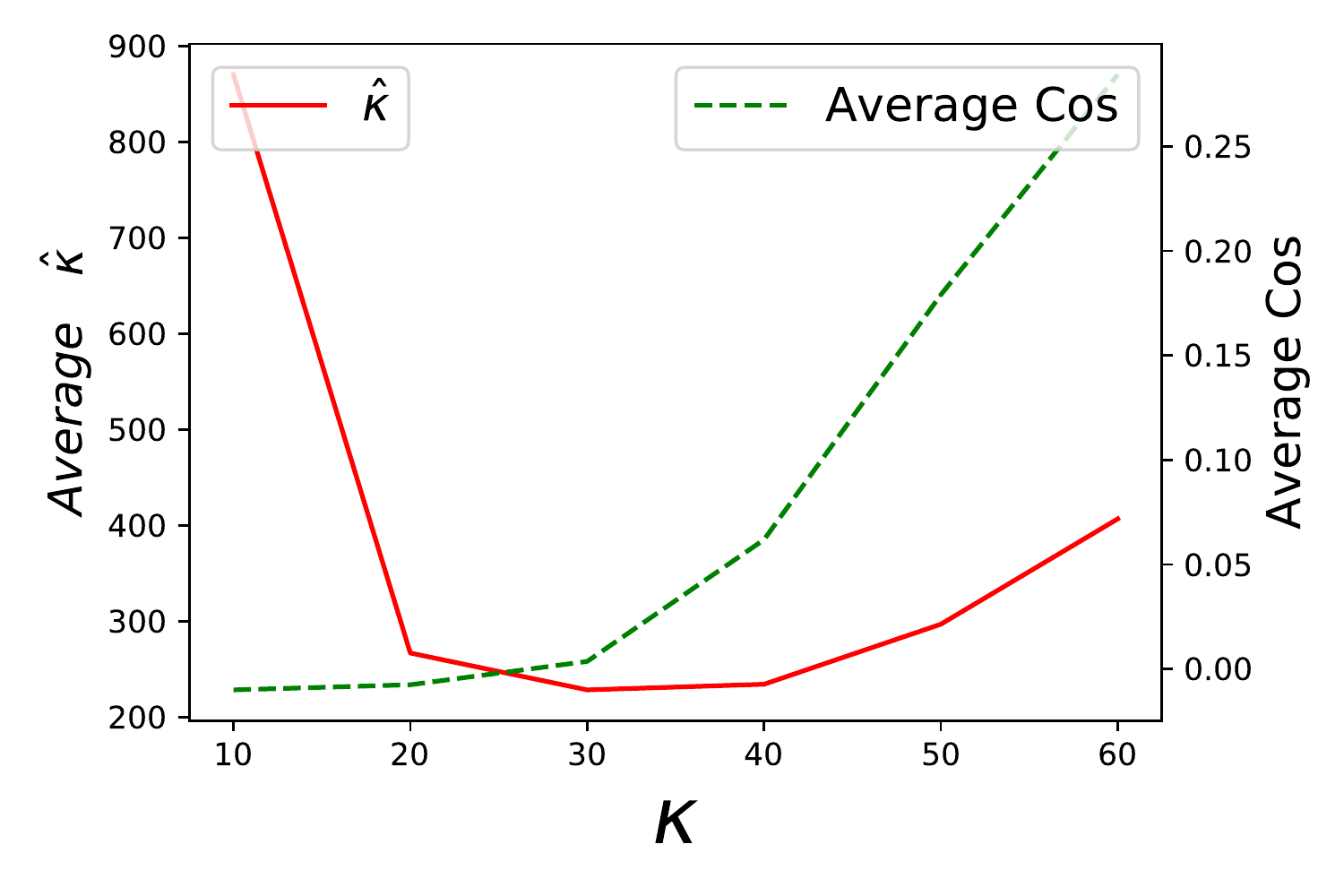}
   \caption{Distribution of training sets in the embedding space.}
\label{fig:distribution}
\end{figure}

To better understand how $\kappa$ influences the vMF model, we calculate the distribution of train data in the learned embedding space based on the \textbf{Average $\hat{\bm{\kappa}}$} and the \textbf{Average Cosine} value between class mean directions.
The $\hat{\kappa}$ here is the true concentration parameter for each class of training data in the embedding space defined in Equation \ref{eq:kappa}.  
The larger value of $\hat{\kappa}$, the higher the concentration of the distribution around the class mean direction. 
A smaller average $\hat{\kappa}$ suggests that the training samples are more evenly distributed among the hyper sphere.
The average cosine value between mean directions of training data is used to represent the class gap.
The smaller average cosine value indicates that the class mean directions are far away from each others.
The results are presented in Figure \ref{fig:distribution}.
It shows that the average cosine between class mean directions is proportional to the hyper parameter $\kappa$.
It indicates that the class gap is smaller when $\kappa$ is greater.
However, too large or too small value of $\kappa$ makes the training data being tightly concentrated around their mean directions.
The smallest average $\hat{\kappa}$ is reached by $\kappa$ around $30\sim40$.
A comparison of the curves of the average $\hat{\kappa}$ and R@1 shows that R@$1$ is roughly inversely proportional to the average $\hat{\kappa}$.
Theoretical discussion of the relationship between the distribution of training data in the embedding space and the retrieval performance is beyond the scope of this paper. 
We give following hypothesis and leave the analysis to future work.
Different from classification task, the classes in testing data for the retrieval task are not included in the training set.
To achieve better retrieval performance, the DDML models should have better global generalization ability.
In order to achieve better global generalization ability, training points should not tightly concentrate around the mean directions.
So a small value of $\hat{\kappa}$ usually has a better retrieval performance.


\section{Conclusion}
\label{sec:conclusion}
In this work, we introduce directional statistics to deep metric learning. 
By considering the feature space as a directional statistical probability space, we propose a new deep metric learning approach.
Specifically, a novel loss function named von Mises-Fisher loss is proposed based on the von Mises-Fisher distribution. 
Then an alternative learning algorithm is applied to train a neural network efficiently with our vMF loss. 
Extensive experiments shows that our vMF can surpass other state-of-the-art methods on both classification and retrieval tasks.

For the future work, we will provide a more analytic method to decide the hyper-parameters in our model. 
As the proposed method can effectively handle the fine-grained categorization problem, applying our model to handle one-shot or zero-shot learning problem will be the top concern for our future work.
\section*{Acknowledgment}
This work is supported by the Hong Kong Research Grants Council (Project C1007-15G), City University of Hong Kong (Project 9610034) and Shenzhen Science and Technology Innovation Committee (Project JCYJ20150401145529049). 
\ifCLASSOPTIONcaptionsoff
  \newpage
\fi



%
\bibliographystyle{IEEEtran}
\bibliography{vmf}

%

\begin{IEEEbiography}[{\includegraphics[width=1in,height=1.25in,clip,keepaspectratio]{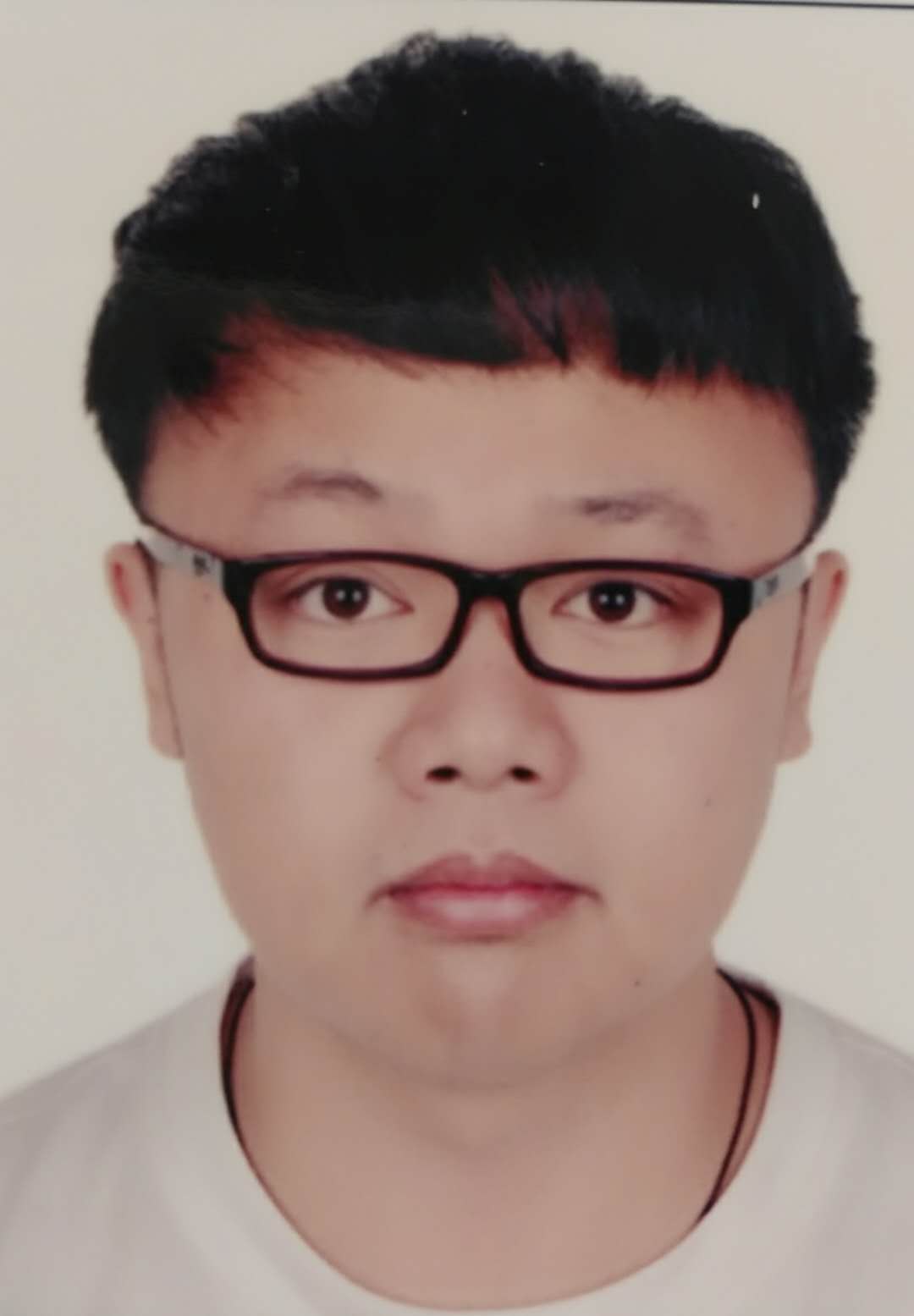}}]{\textbf{Xuefei Zhe} received the B.Sc. degree in Information Engineering from Nanjing University, China in 2014. He is currently working toward the Ph.D. degree in the Department of Electronic Engineering, City University of Hong Kong. His research interests include computer vision and deep learning.}
\end{IEEEbiography}


\begin{IEEEbiography}[{\includegraphics[width=1in,height=1.25in,clip,keepaspectratio]{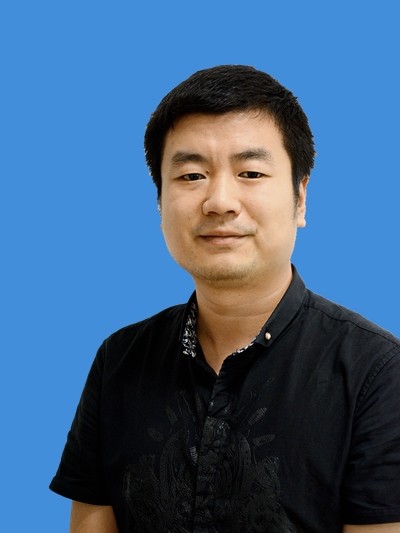}}]{\textbf{Shifeng Chen} received the B.E. degree from the University of Science and Technology of China, Hefei, in 2002, the M.Phil. degree from City University of Hong Kong, Hong Kong, in 2005, and the Ph.D. Degree from the Chinese University of Hong Kong, Hong Kong, in 2008. He is now an Associate Professor in the Shenzhen Institutes of Advanced Technology, Chinese Academy of Sciences, China. His research interests include computer vision and machine learning.}
\end{IEEEbiography}

\begin{IEEEbiography}[{\includegraphics[width=1in,height=1.25in,clip,keepaspectratio]{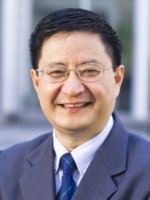}}]{\textbf{Hong Yan} received his Ph.D. degree from Yale University. He was professor of imaging science at the University of Sydney and currently is professor of computer engineering at City University of Hong Kong. His research interests include image processing, pattern recognition and bioinformatics. He has authored or co-authored over 300 journal and conference papers in these areas. He was elected an IAPR fellow for contributions to document image analysis and an IEEE fellow for contributions to image recognition techniques and applications. He received the 2016 Norbert Wiener Award from IEEE SMC Society for contributions to image and biomolecular pattern recognition techniques.}
\end{IEEEbiography}




\end{document}